\documentclass{article}

\usepackage{wrapfig}
\usepackage{stfloats}
\usepackage{graphicx}
\usepackage{subcaption}
\usepackage[preprint]{neurips_2026}
\usepackage[space]{grffile}
\usepackage[utf8]{inputenc} 
\usepackage[T1]{fontenc}          
\usepackage{url}            
\usepackage{booktabs}       
\usepackage{amsfonts}       
\usepackage{nicefrac}       
\usepackage{microtype}      
\usepackage{xcolor}         
\usepackage{graphicx}       
\usepackage{amsmath}        
\usepackage{amssymb}
\usepackage{multirow}
\usepackage{pifont}
\usepackage{booktabs}   
\usepackage{multirow}   
\usepackage{graphicx}   
\usepackage{pifont}     
\usepackage[table]{xcolor} 
\usepackage{tabularx}
\usepackage{enumitem}
\usepackage{adjustbox}

\definecolor{highlightpink}{RGB}{255, 230, 230}
\definecolor{myyellow}{RGB}{255, 255, 200}
\definecolor{myblue}{RGB}{210, 240, 255}
\definecolor{mypink}{RGB}{255, 230, 240}
\definecolor{lightblue}{RGB}{51, 153, 255} 
\definecolor{winered}{RGB}{128, 0, 0}      
\definecolor{deepmaroon}{RGB}{153, 0, 51}
\usepackage[colorlinks=true, linkcolor=deepmaroon, citecolor=lightblue]{hyperref}
\newcommand{\greenCheck}{{\color{green!60!black}\ding{51}}}
\newcommand{\redCross}{{\color{red}\ding{55}}}
\title{AHPA: Adaptive Hierarchical Prior Alignment for Diffusion Transformers}
\author{%
  \textbf{Ruibin Min}$^{1,2}$, 
  \textbf{Yexin Liu}$^{2, \ast}$, 
  \textbf{Aimin Pan}$^{3}$, 
  \textbf{Changsheng Lu}$^{2}$, \\
  \textbf{Jiafei Wu}$^{5}$, 
  \textbf{Kelu Yao}$^{4}$, 
  \textbf{Xiaogang Xu}$^{5}$, 
  \textbf{Harry Yang}$^{2, \dagger}$ \\
  \vspace{1mm} \\
  $^{1}$Sun Yat-sen University \quad $^{2}$The Hong Kong University of Science and Technology \\
  $^{3}$VNET Group \quad $^{4}$Zhejiang Lab \quad
  $^{5}$Zhejiang University \\
  $^\ast$Project Leader \quad $^\dagger$Corresponding Author \\
  \texttt{minrb@mail2.sysu.edu.cn} \\
}
\begin{document}

\maketitle
\vspace{-2.5em}
\begin{figure*}[ht]
    \centering
    \includegraphics[width=0.1666\textwidth]{"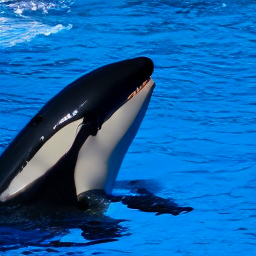"}%
    \includegraphics[width=0.1666\textwidth]{"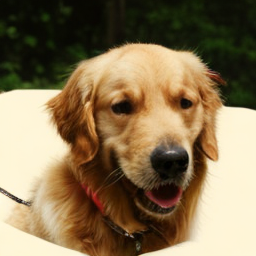"}%
    \includegraphics[width=0.1666\textwidth]{"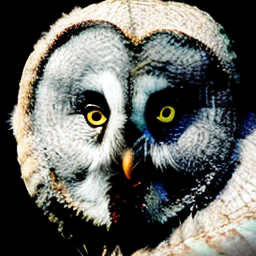"}%
    \includegraphics[width=0.1666\textwidth]{"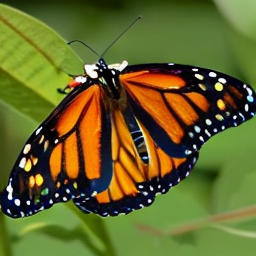"}%
    \includegraphics[width=0.1666\textwidth]{"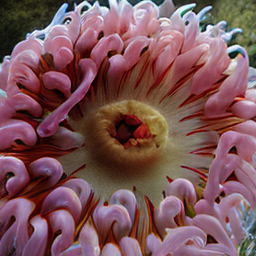"}%
    \includegraphics[width=0.1666\textwidth]{"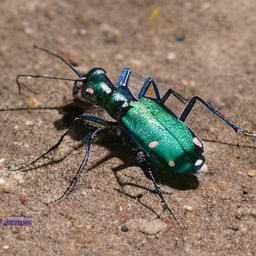"}\\[-1pt] 
    \nointerlineskip 
    \includegraphics[width=0.1666\textwidth]{"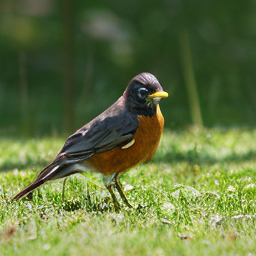"}%
    \includegraphics[width=0.1666\textwidth]{"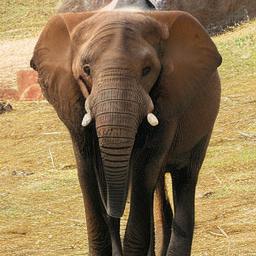"}%
    \includegraphics[width=0.1666\textwidth]{"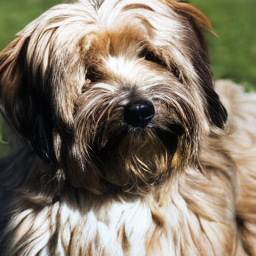"}%
    \includegraphics[width=0.1666\textwidth]{"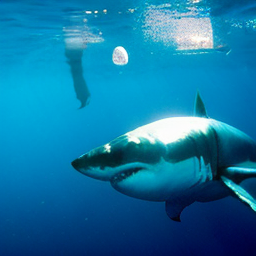"}%
    \includegraphics[width=0.1666\textwidth]{"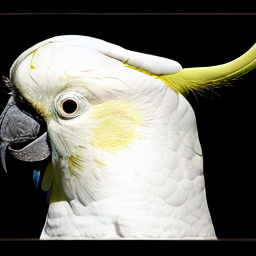"}%
    \includegraphics[width=0.1666\textwidth]{"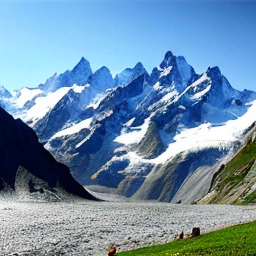"}\\[-1pt]
    \nointerlineskip
    \includegraphics[width=0.1666\textwidth]{"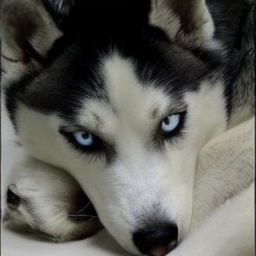"}%
    \includegraphics[width=0.1666\textwidth]{"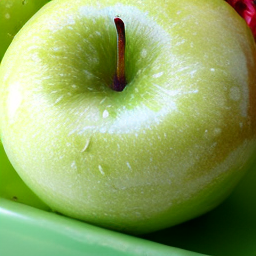"}%
    \includegraphics[width=0.1666\textwidth]{"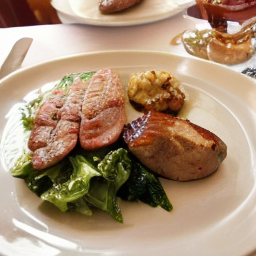"}%
    \includegraphics[width=0.1666\textwidth]{"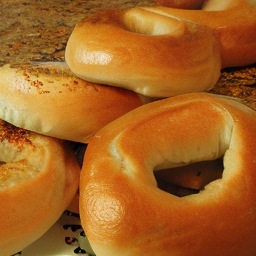"}%
    \includegraphics[width=0.1666\textwidth]{"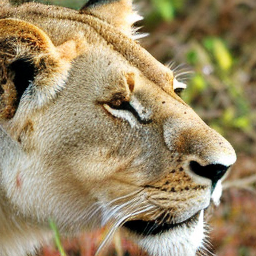"}%
    \includegraphics[width=0.1666\textwidth]{"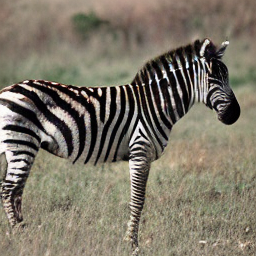"}

    \caption{\textbf{High-fidelity image generated by SiT-XL/2 with AHPA alignment.}}
    \label{fig:main_results}
\end{figure*}
\begin{abstract}

Representation alignment has recently emerged as an effective paradigm for accelerating Diffusion Transformer training. Despite their success, existing alignment methods typically impose a fixed supervision target or a fixed alignment granularity throughout the entire denoising trajectory, whether the guidance is provided by external vision encoders, internal self-representations, or VAE-derived features. We argue that such timestep-agnostic alignment is suboptimal because the useful granularity of representation supervision changes systematically with the signal-to-noise ratio. In high-noise regimes, diffusion models benefit more from coarse semantic and layout-level anchoring, whereas in low-noise regimes, the training signal should emphasize spatially detailed and structurally faithful refinement. 
This non-stationary alignment behavior creates a representational mismatch for static single-level supervisors. To address this issue, we propose \textbf{Adaptive Hierarchical Prior Alignment (AHPA)}, a lightweight alignment framework that exploits the hierarchical representations naturally embedded in the frozen VAE encoder. 
Instead of using only a single compressed latent as the alignment target, AHPA extracts multi-level VAE features that provide complementary priors ranging from local geometry and spatial topology to coarse semantic layout. 
A timestep-conditioned Dynamic Router adaptively selects and weights these hierarchical priors along the denoising trajectory, thereby synchronizing the alignment granularity with the model's evolving training needs. 
Extensive experiments show that AHPA improves convergence and generation quality over baselines and incurs no additional inference cost while avoiding external encoder supervision during training. Our code is available at \url{https://github.com/UIOSN/AHPA}.
\end{abstract}

\section{Introduction}

Recent advances in diffusion models~\cite{ho2020ddpm,song2021denoising} have substantially improved high-fidelity image generation, with Diffusion Transformers~\cite{peebles2023scalable,ma2024sit} emerging as a scalable backbone for generative modeling. 
However, training such models from scratch remains computationally demanding, often requiring long optimization schedules before the denoising network learns semantically meaningful and structurally coherent representations. 
Representation alignment has recently been introduced as an effective paradigm for accelerating this process: by aligning intermediate transformer features with informative visual representations, the denoising backbone can acquire stronger semantic and structural priors during training~\cite{yu2024repa,singh2025irepa,wu2025reg,chen2025sara}. For example, REPA-style methods transfer knowledge from external vision encoders such as DINOv2~\cite{oquab2024dinov2learningrobustvisual}, while recent variants further improve the alignment objective through structural constraints, adversarial matching, or improved representation regularization~\cite{singh2025irepa,chen2025sara}.

Despite these advances, existing alignment strategies are still largely built upon a fixed-target formulation. 
Most methods differ in the source of supervision---external vision encoders, internal self-representations, or VAE-derived features---but typically impose a fixed type of alignment signal throughout the denoising trajectory. 
External-teacher-based methods provide strong semantic guidance but introduce additional encoder forward passes, memory usage, and training complexity. 
Some alternatives such as SRA~\cite{jiang2026sra} and DUPA~\cite{peng2026dupa} reduce the dependence on external teachers through self-distillation or dual-path consistency, yet may still incur extra training computation. 
Recent VAE-based methods such as SRA2~\cite{wang2026sra2variationalautoencoder} further reduce the teacher cost by using VAE-derived supervision; however, relying on a single compressed latent representation provides limited access to the fine-grained intermediate structures that are important for effective alignment~\cite{yao2025reconstruction,singh2025irepa}.

A more intrinsic limitation lies in the timestep-agnostic nature of existing alignment objectives. 
Diffusion training is inherently non-stationary: as the signal-to-noise ratio evolves, the useful granularity of representation supervision changes accordingly. 
In high-noise regimes, the denoising network benefits from coarse semantic and layout-level anchoring, while in low-noise regimes, it requires spatially detailed and structurally faithful guidance. 
A fixed single-level supervisor therefore cannot remain equally suitable across all timesteps, resulting in a granularity mismatch between the alignment target and the denoising state. 
To analyze this, we introduce the Gradient Signal-to-Noise Ratio (G-SNR, Sec.~\ref{sec:motivation}) to measure the alignment health along the trajectory, which reveals that static methods suffer from declining directional consistency and gradient interference. 
This mismatch limits the acceleration potential of static representation alignment.

Motivated by this observation, we propose \textbf{Adaptive Hierarchical Prior Alignment (AHPA)}, which reformulates representation alignment from static target matching into timestep-adaptive hierarchical prior scheduling. 
Instead of treating the frozen VAE encoder~\cite{rombach2022LDM-4} merely as a latent compression module, AHPA re-examines it as a multi-scale repository of hierarchical priors. 
Intermediate VAE features naturally encode complementary information at different levels, ranging from local geometry and spatial topology to coarse layout-level structures. 
To exploit these priors effectively, AHPA introduces a lightweight timestep-conditioned dynamic router that adaptively selects and weights hierarchical VAE features according to the current denoising stage. 
This design synchronizes the granularity of alignment supervision with the evolving needs of diffusion training, while avoiding heavyweight external teacher encoders and introducing no additional inference-time cost.

Our main contributions are summarized as follows:
\begin{itemize}[leftmargin=*, itemsep=0.2em, topsep=0.2em]
    \item \textbf{Non-stationary Alignment Analysis.} 
    We identify that the effective granularity of representation alignment varies across denoising timesteps, and show that fixed single-level supervision can lead to a representational mismatch during diffusion training.

    \item \textbf{Hierarchical VAE Prior Mining.} 
    We reveal that intermediate layers of the frozen VAE encoder provide a natural multi-scale prior bank, offering complementary guidance from local structural details to coarse layout-level representations.

    \item \textbf{Adaptive and Efficient Alignment.} 
    We propose AHPA, a timestep-aware alignment framework equipped with a lightweight dynamic router, achieving adaptive hierarchical supervision with zero additional inference cost and reduced dependence on heavyweight external teachers.
\end{itemize}
\section{Related Work}

\paragraph{Generative Models and Diffusion Transformers.}
Image generation has witnessed a transition from pixel-space denoising models, such as DDPM~\cite{ho2020ddpm} and DDIM~\cite{song2021denoising}, to Latent Diffusion Models (LDMs)~\cite{rombach2022LDM-4} that operate in compressed latent spaces. Architecturally, while early frameworks relied on U-Net backbones~\cite{ho2020ddpm,rombach2022LDM-4}, modern paradigms have shifted towards Transformer-based architectures like DiT~\cite{peebles2023scalable} and SiT~\cite{ma2024sit}, which leverage self-attention mechanisms for superior spatial modeling~\cite{peebles2023scalable}. Despite these advances, diffusion transformers typically require extensive training iterations to achieve convergence~\cite{singh2025irepa,wu2025reg}.

\noindent \textbf{Representation Alignment and Entanglement.}
A prominent direction to expedite training is leveraging discriminative priors from pretrained vision encoders. REPA~\cite{yu2024repa} pioneered this by aligning intermediate diffusion features with representations from external teachers like DINOv2~\cite{oquab2024dinov2learningrobustvisual}. Building on this, iREPA~\cite{singh2025irepa} and SARA~\cite{chen2025sara} further enhanced alignment quality by focusing on spatial structures or introducing adversarial distribution matching. While effective, these methods are teacher-dependent, requiring external models that introduce significant memory overhead and inherent domain biases~\cite{singh2025irepa, wu2025reg}. Alternatively, REG~\cite{wu2025reg} proposes entangling high-level class tokens directly into the diffusion process for semantic guidance, yet still relies on fixed category embeddings.

\noindent \textbf{Self-Supervised Alignment and its Bottlenecks.}
To eliminate dependence on external teachers, a new frontier of autonomous alignment has emerged. SRA~\cite{jiang2026sra} and SRA 2~\cite{wang2026sra2variationalautoencoder} propose self-guidance by distilling features from either an EMA-updated teacher model or the final output of a frozen VAE encoder. Similarly, DUPA~\cite{peng2026dupa} achieves self-alignment by enforcing consistency between dual noising paths of the same image. However, they rely on static guidance, utilizing a monolithic feature target throughout the entire process. Such "time-invariant" strategies ignore the non-stationary alignment requirement inherent in diffusion models. AHPA fills this gap by adaptively routing hierarchical VAE priors, ensuring the guidance granularity is synchronized with the model's shifting receptivity along the denoising trajectory.

\begin{figure*}[t]
    \centering
    \begin{subfigure}[b]{0.32\textwidth}
        \centering
        \includegraphics[width=\textwidth]{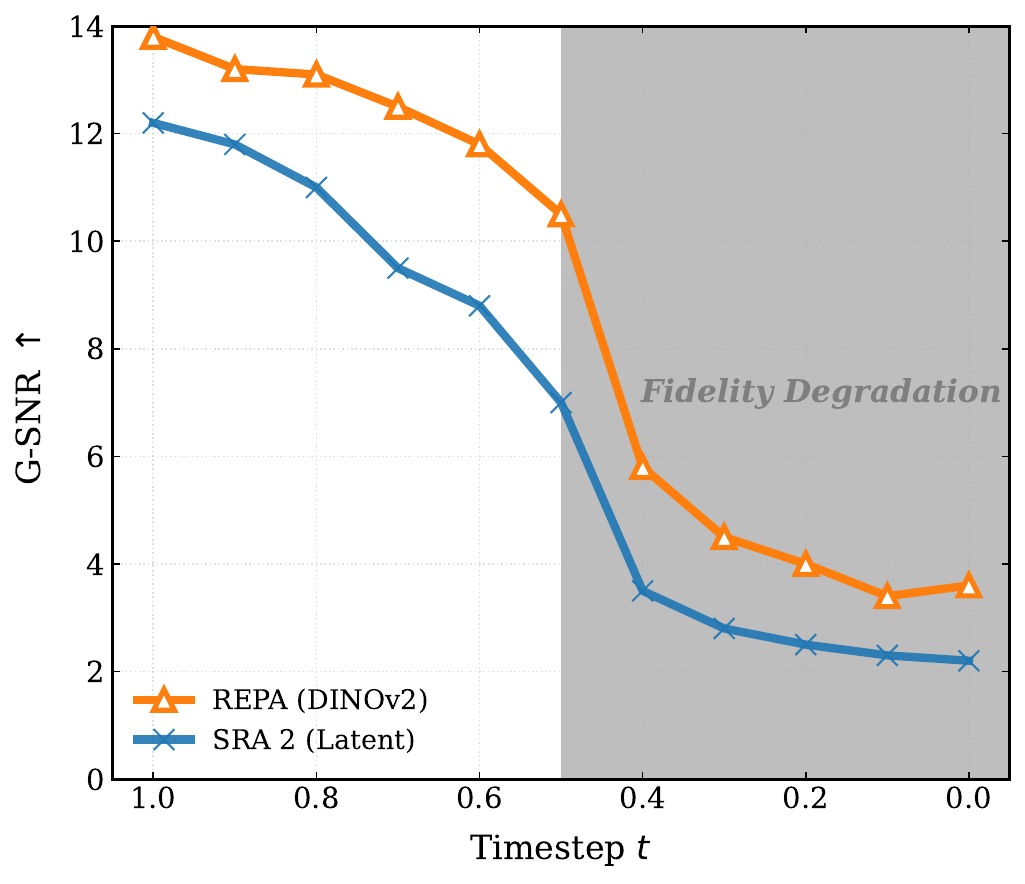}
        \caption{Diagnostic Probing}
        \label{fig:2a}
    \end{subfigure}%
    \hfill
    \begin{subfigure}[b]{0.32\textwidth}
        \centering
        \includegraphics[width=\textwidth]{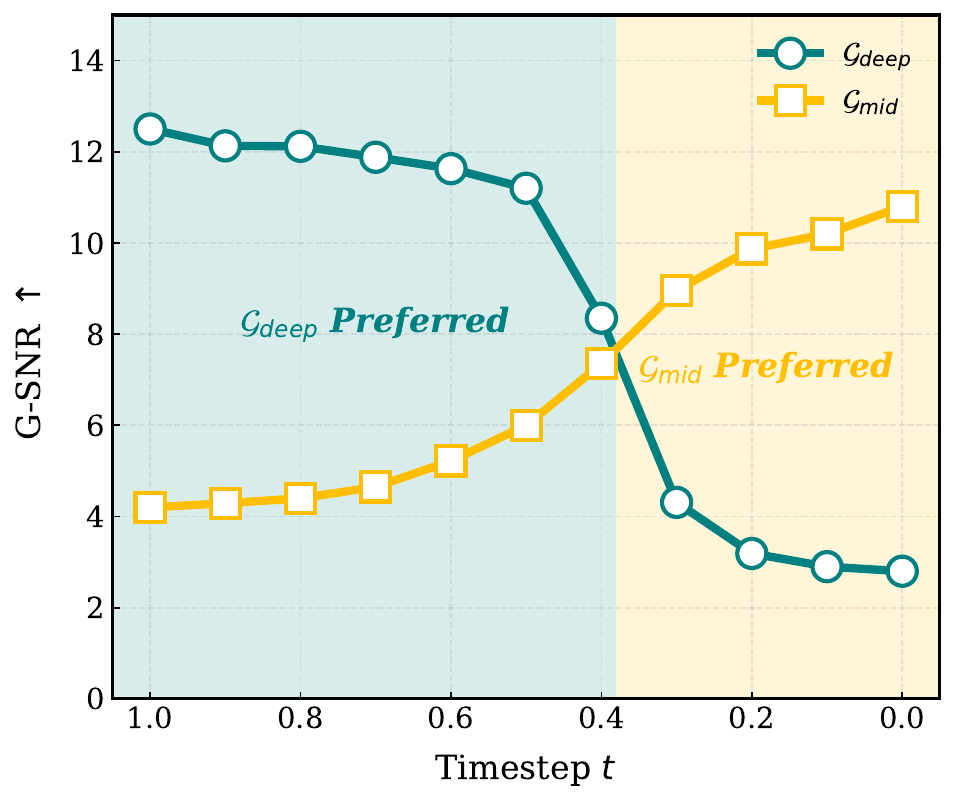}
        \caption{Complementarity Discovery}
        \label{fig:2b}
    \end{subfigure}%
    \hfill
    \begin{subfigure}[b]{0.32\textwidth}
        \centering
        \includegraphics[width=\textwidth]{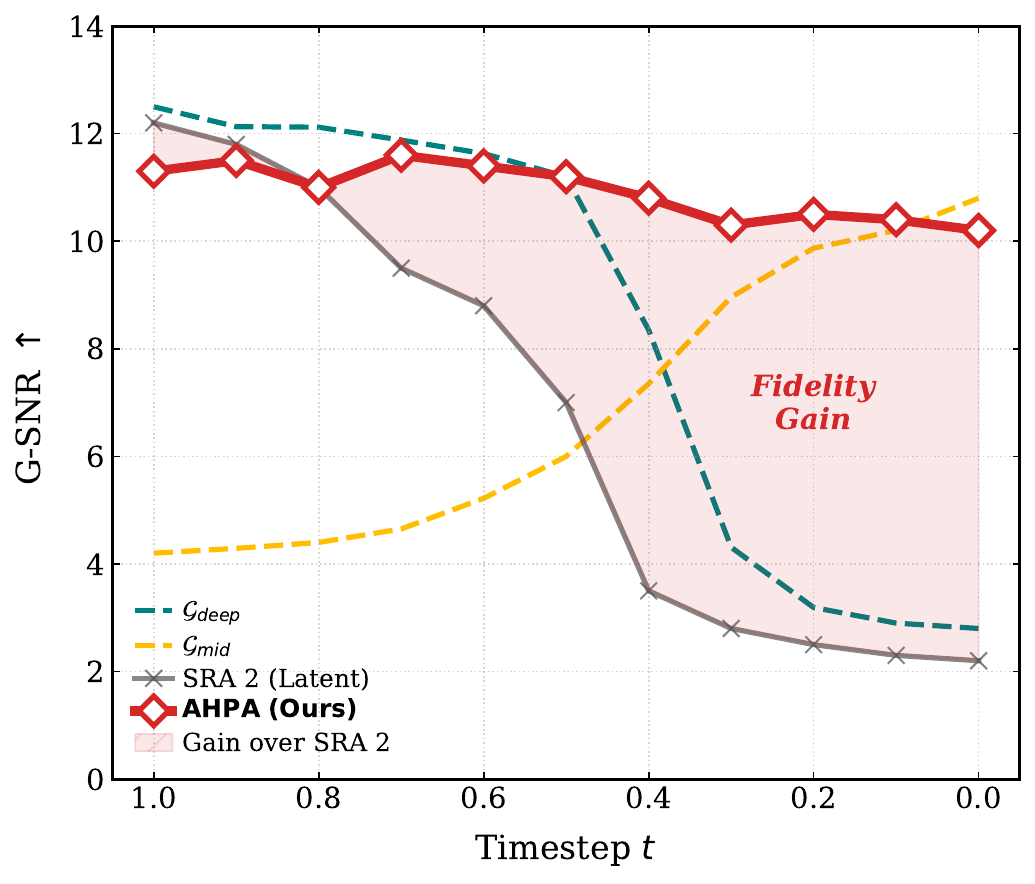}
        \caption{Proposal \& Verification}
        \label{fig:2c}
    \end{subfigure}
    \caption{\textbf{Quantifying the non-stationary alignment requirement.} (a) Diagnostic probing reveals that static baselines suffer from catastrophic fidelity degradation as $t \to 0$. (b) Hierarchical features ($\mathcal{G}_{deep}$ and $\mathcal{G}_{mid}$) exhibit significant phase complementarity. (c) Our AHPA adaptively bridges these gaps to maintain a high-level G-SNR envelope throughout the full trajectory.}
    \label{fig:motivation_main}
\vspace{-12pt}
\end{figure*}

\section{Motivation: Identifying and Resolving Representational Mismatch}
\label{sec:motivation}

It is widely recognized that the generative trajectory of diffusion models is inherently multi-stage \cite{ho2020ddpm,nichol2021improved,choi2022perception}, typically transitioning from global structural anchoring in high-noise regimes to local texture refinement as $t \to 0$. This progressive nature raises a critical question: \textit{can a static supervisor effectively guide a non-stationary denoising process?} We hypothesize that a static approach, by utilizing a fixed-granularity prior, inevitably creates a mismatch with the model's evolving representational needs. To rigorously investigate this mismatch and its impact on alignment health, we introduce the \textbf{Gradient Signal-to-Noise Ratio (G-SNR)}. Let $g_t = \nabla_{\theta} \mathcal{L}_{align}$ be the gradient elicited by the alignment objective at timestep $t$; we define:
\begin{equation}
\footnotesize
    \text{G-SNR}(t) = \frac{\| \mathbb{E}[g_t] \|_2^2}{\text{Tr}(\text{Var}(g_t))}
\end{equation}
G-SNR measures signal purity: a high G-SNR indicates directionally consistent guidance across the mini-batch, whereas a collapse implies severe gradient interference. To provide a reliable estimate, we compute $\text{G-SNR}(t)$ using the variance across per-sample gradients within each mini-batch. This formulation renders the metric dimensionless and invariant to absolute gradient magnitudes, allowing for a consistent assessment of alignment quality across different timesteps and layers. A formal discussion on the variance estimator and its theoretical grounding is provided in Appendix ~\ref{app:metric_validation}.

\noindent\textbf{Diagnostic Probing: The Collapse of Static Baselines.} 
We first quantify the gradient properties of existing acceleration paradigms, such as REPA \cite{yu2024repa} and SRA 2 \cite{wang2026sra2variationalautoencoder}. As illustrated in Fig.~\ref{fig:2a}, while these static methods perform well in early stages, their G-SNR undergoes a dramatic collapse as $t \to 0$, indicating that the guidance becomes directionally inconsistent and suffers from gradient interference. This degradation reveals a bottleneck: a single, fixed representational granularity cannot adapt to the non-stationary SNR environment throughout the denoising trajectory. Consequently, we posit that the key to sustained acceleration lies in a multi-scale feature library that can dynamically match the model's evolving requirements across different phases. This motivates us to explore the internal hierarchy of the VAE encoder, leveraging its multi-scale feature space as a natural source of adaptive 'blueprints' that correspond to different denoising stages. 

\noindent\textbf{The Complementarity Discovery: Phase-wise Synergies.} 
We perform a full-spectrum probe of the VAE encoder's hierarchical features, Our analysis reveals that while the shallow layers consistently yield low G-SNR due to excessive pixel-level noise and a lack of semantic coherence, a significant phase complementarity emerges between the mid-level ($\mathcal{G}_{mid}$) and deep-level ($\mathcal{G}_{deep}$) features. As illustrated in Fig.~\ref{fig:2b}, $\mathcal{G}_{deep}$ maintains higher directional consistency during the high-noise regimes ($t > 0.4$), whereas $\mathcal{G}_{mid}$ takes over as the more reliable guide for structural details as $t \to 0$. This crossover confirms that while no single layer can dominate the entire trajectory, their synergy offers a potential path to sustaining high-purity guidance.

\noindent\textbf{Proposal \& Verification: AHPA Trajectory Optimization.} 
Driven by these insights, we propose AHPA, which utilizes a dynamic router to switch or fuse these complementary features in real-time based on $t$. Experimental verification (Fig.~\ref{fig:2c}) confirms that this adaptive scheduling successfully maintains a high-level G-SNR envelope across the full trajectory. This global enhancement in gradient quality directly translates into faster manifold convergence and superior image generation quality, effectively capturing the fidelity gain previously lost to static methods.

\begin{figure}[t]
    \centering
    \includegraphics[width=\linewidth]{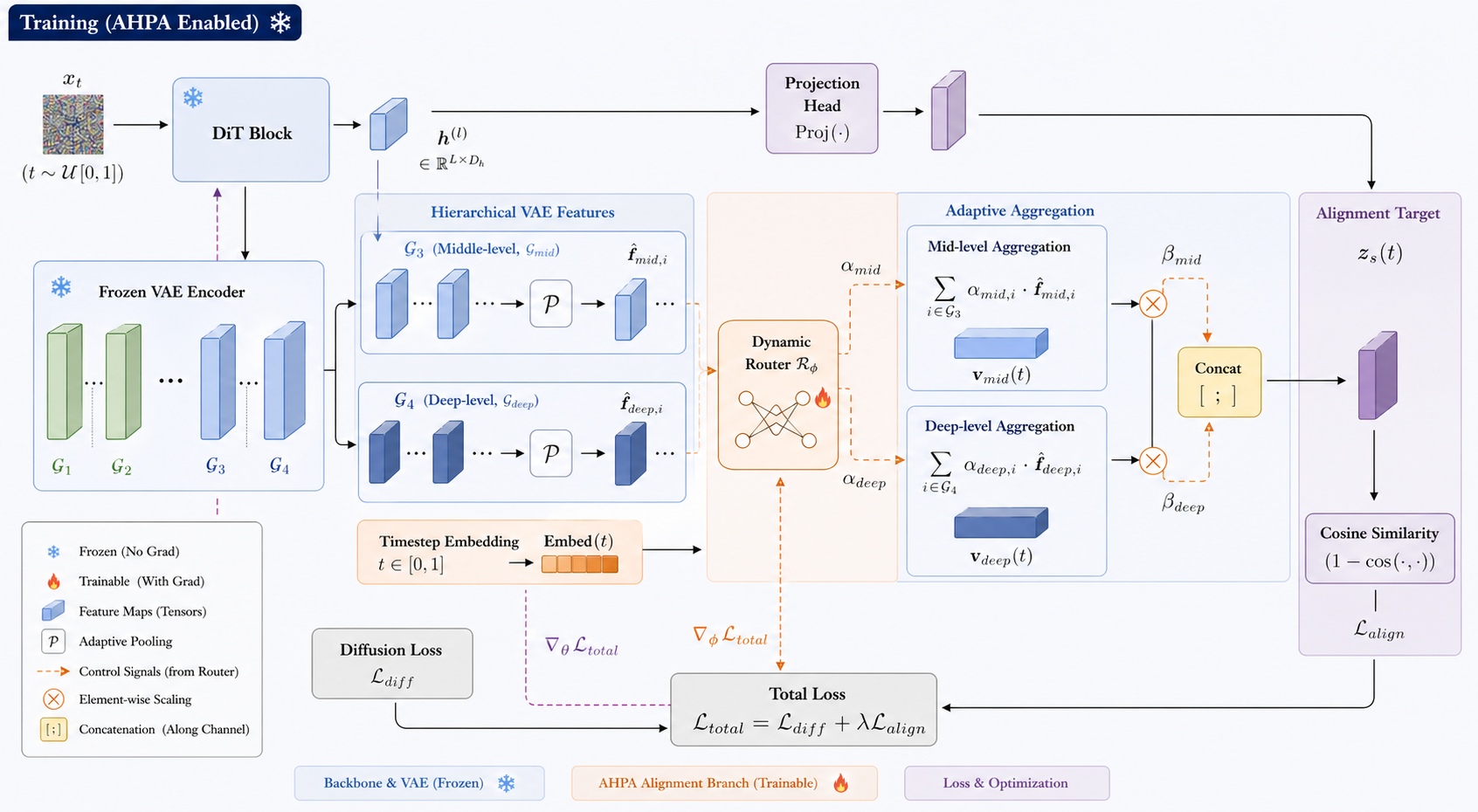}
    \caption{\textbf{Overview of AHPA.} AHPA extracts multi-scale hierarchical priors from a frozen VAE encoder. A timestep-conditioned dynamic router $\mathcal{R}_\phi$ adaptively schedules these priors ($\alpha, \beta$) to align with the DiT backbone's evolving needs, incurring zero inference overhead.}
    \label{fig:pipeline}
    \vspace{-10pt}
\end{figure}

\section{Method}

\subsection{Adaptive Hierarchical Prior Alignment (AHPA)}
The core of AHPA is that the diffusion process is inherently non-stationary, requiring a dynamic shift in guidance granularity as the signal-to-noise ratio (SNR) evolves. We propose to align the internal representations of the Transformer with a multi-scale features extracted from a frozen VAE encoder, ensuring the guidance remains synchronized with the model's time-varying representational needs. The overall pipeline of AHPA is visualized in Fig.~\ref{fig:pipeline}.

\subsection{Mining Hierarchical Representation Blueprints}
\label{Mining}
\begin{figure*}[t]
    \centering
    \begin{subfigure}[b]{0.142\textwidth} \centering \includegraphics[width=\textwidth]{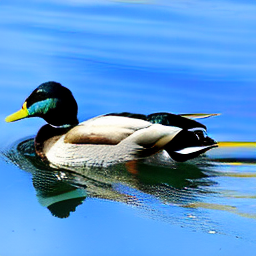} \end{subfigure}%
    \begin{subfigure}[b]{0.142\textwidth} \centering \includegraphics[width=\textwidth]{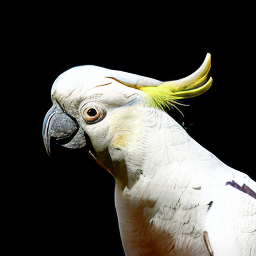} \end{subfigure}%
    \begin{subfigure}[b]{0.142\textwidth} \centering \includegraphics[width=\textwidth]{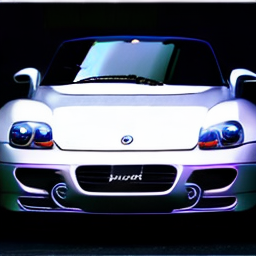} \end{subfigure}%
    \begin{subfigure}[b]{0.142\textwidth} \centering \includegraphics[width=\textwidth]{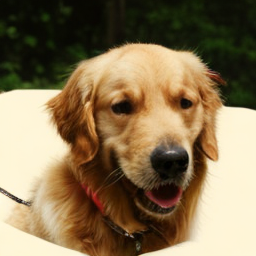} \end{subfigure}%
    \begin{subfigure}[b]{0.142\textwidth} \centering \includegraphics[width=\textwidth]{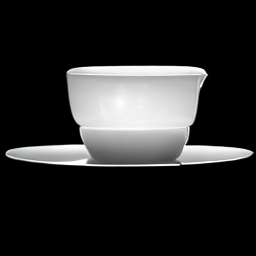} \end{subfigure}%
    \begin{subfigure}[b]{0.142\textwidth} \centering \includegraphics[width=\textwidth]{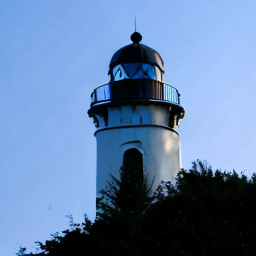} \end{subfigure}%
    \begin{subfigure}[b]{0.142\textwidth} \centering \includegraphics[width=\textwidth]{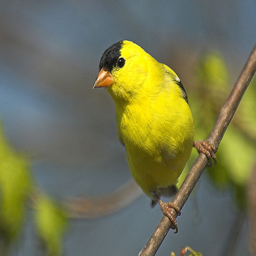} \end{subfigure} \\
    \vspace{-1pt} 
    
    \begin{subfigure}[b]{0.142\textwidth} \centering \includegraphics[width=\textwidth]{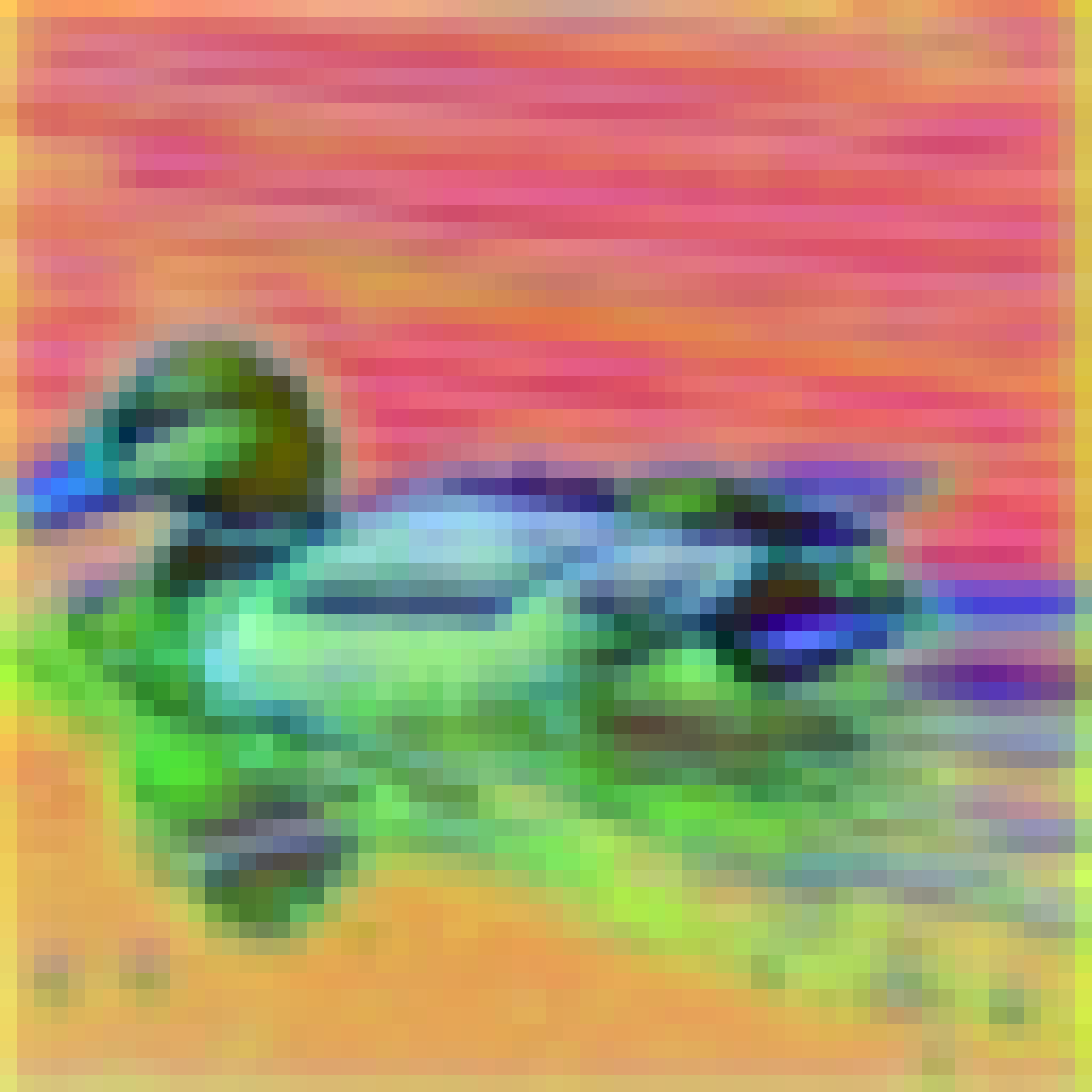} \end{subfigure}%
    \begin{subfigure}[b]{0.142\textwidth} \centering \includegraphics[width=\textwidth]{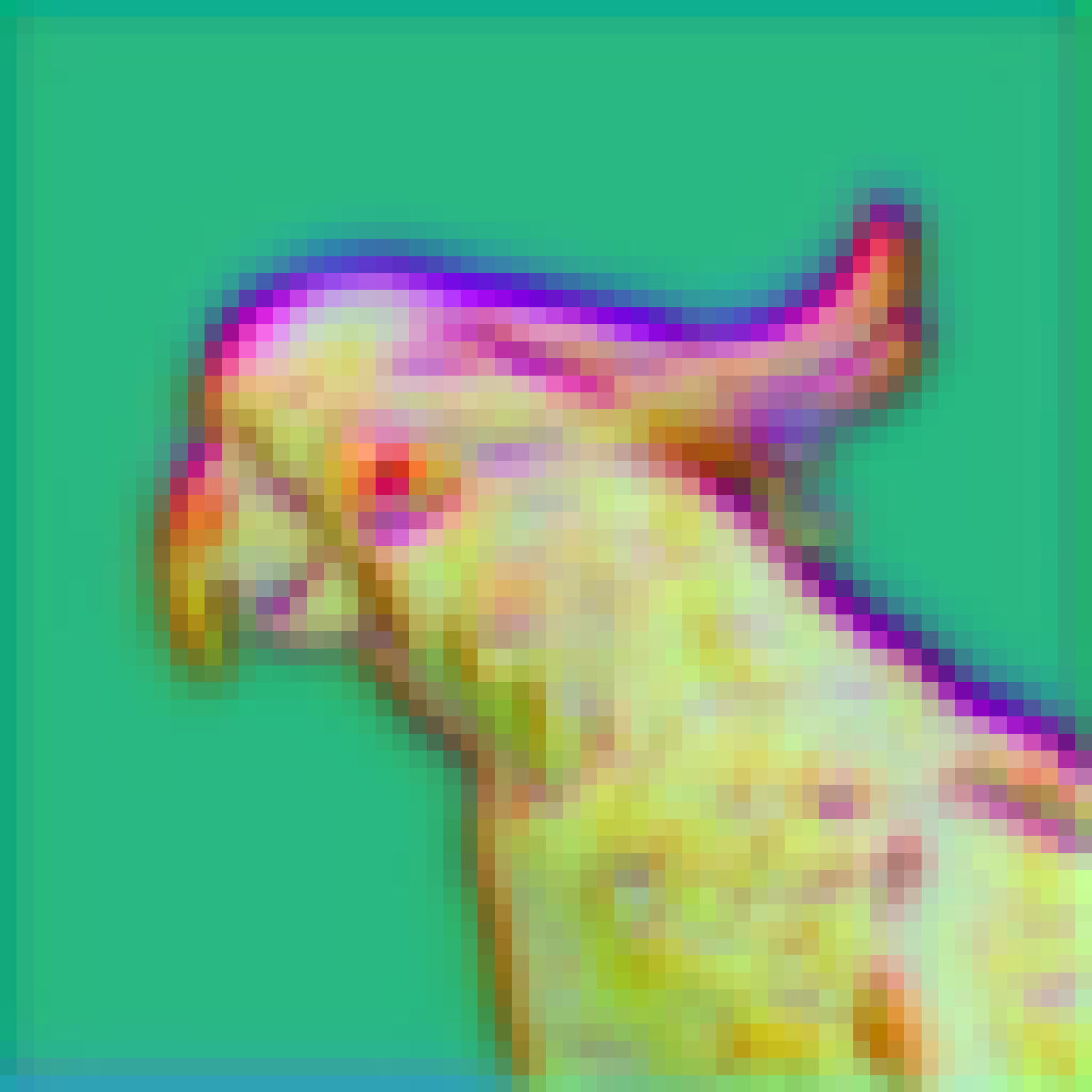} \end{subfigure}%
    \begin{subfigure}[b]{0.142\textwidth} \centering \includegraphics[width=\textwidth]{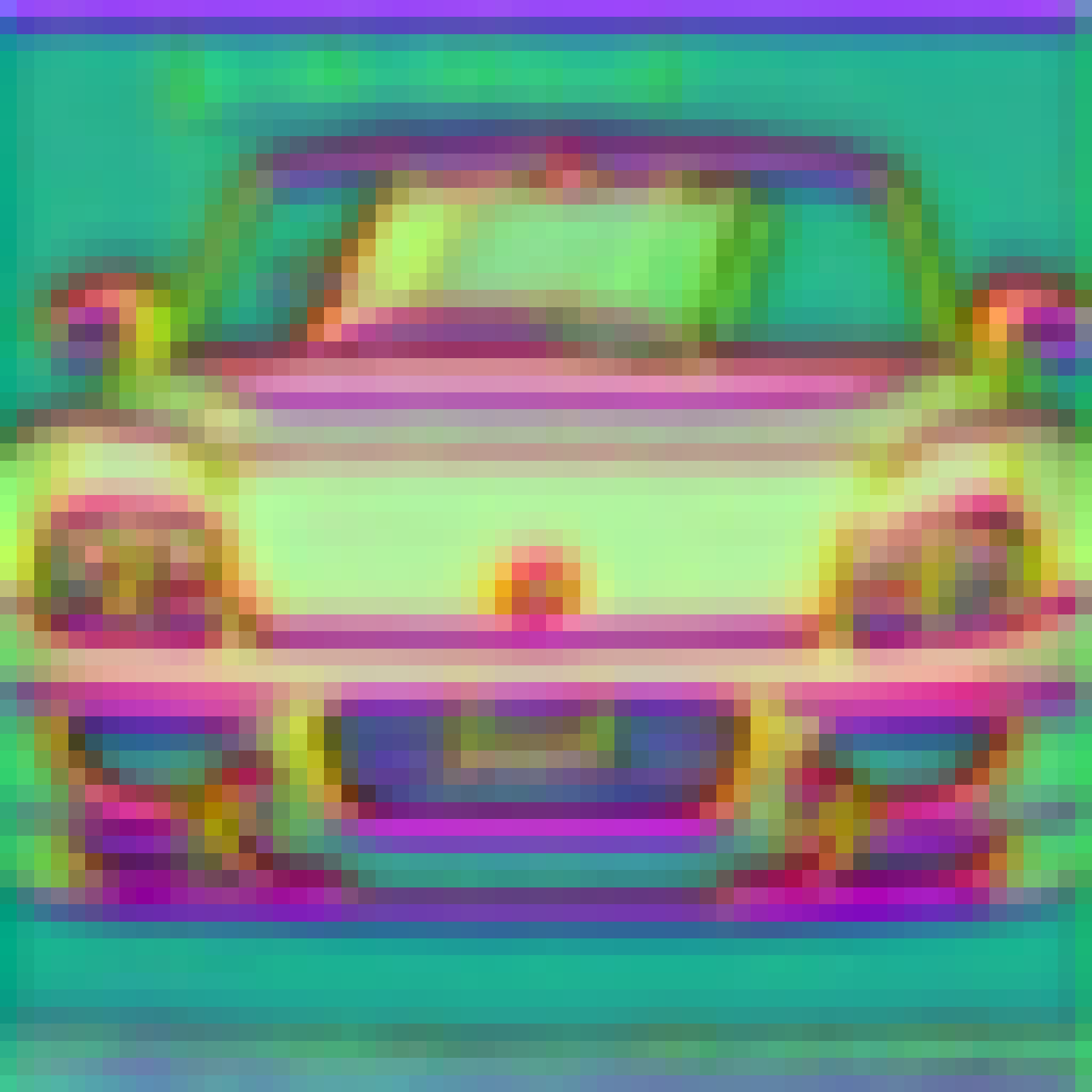} \end{subfigure}%
    \begin{subfigure}[b]{0.142\textwidth} \centering \includegraphics[width=\textwidth]{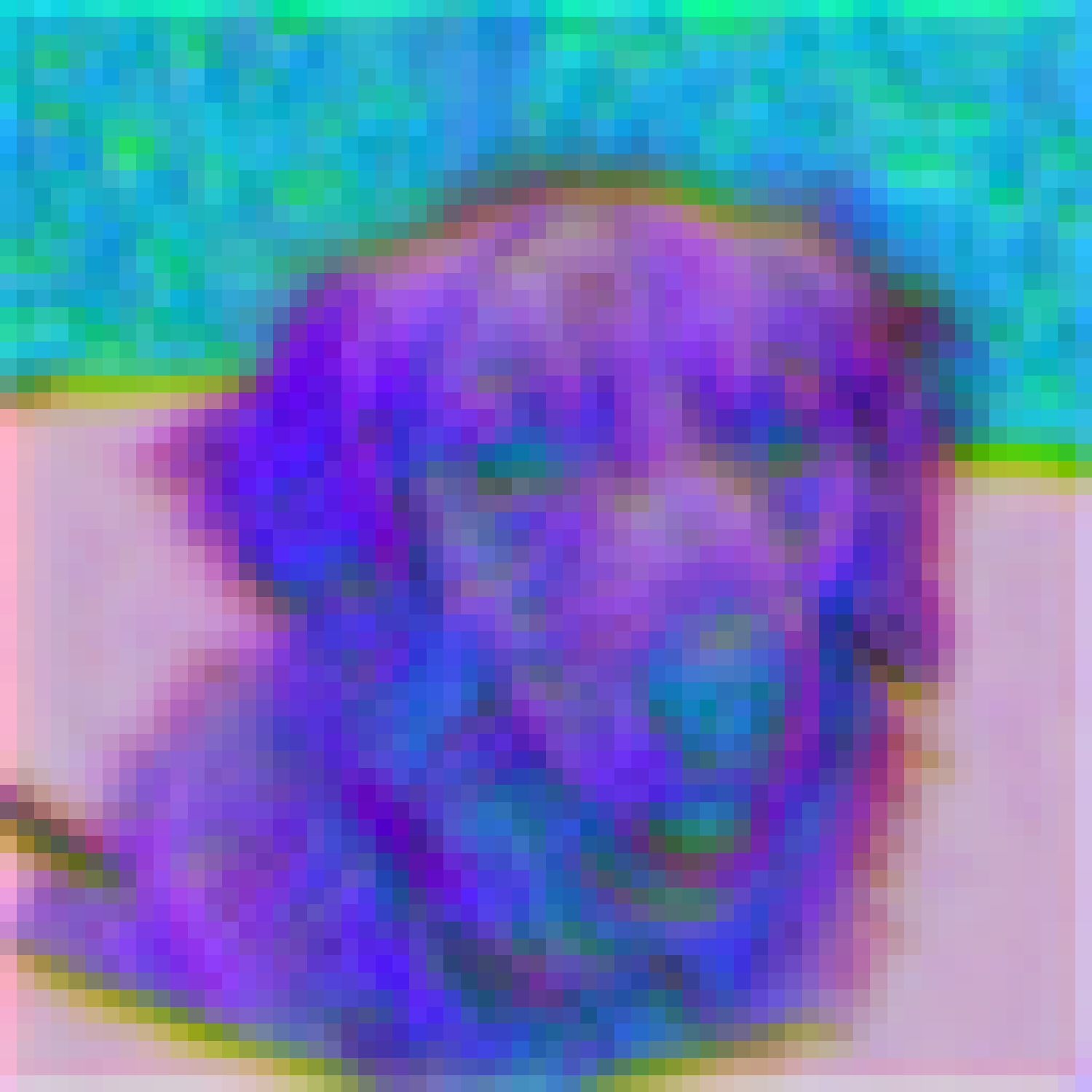} \end{subfigure}%
    \begin{subfigure}[b]{0.142\textwidth} \centering \includegraphics[width=\textwidth]{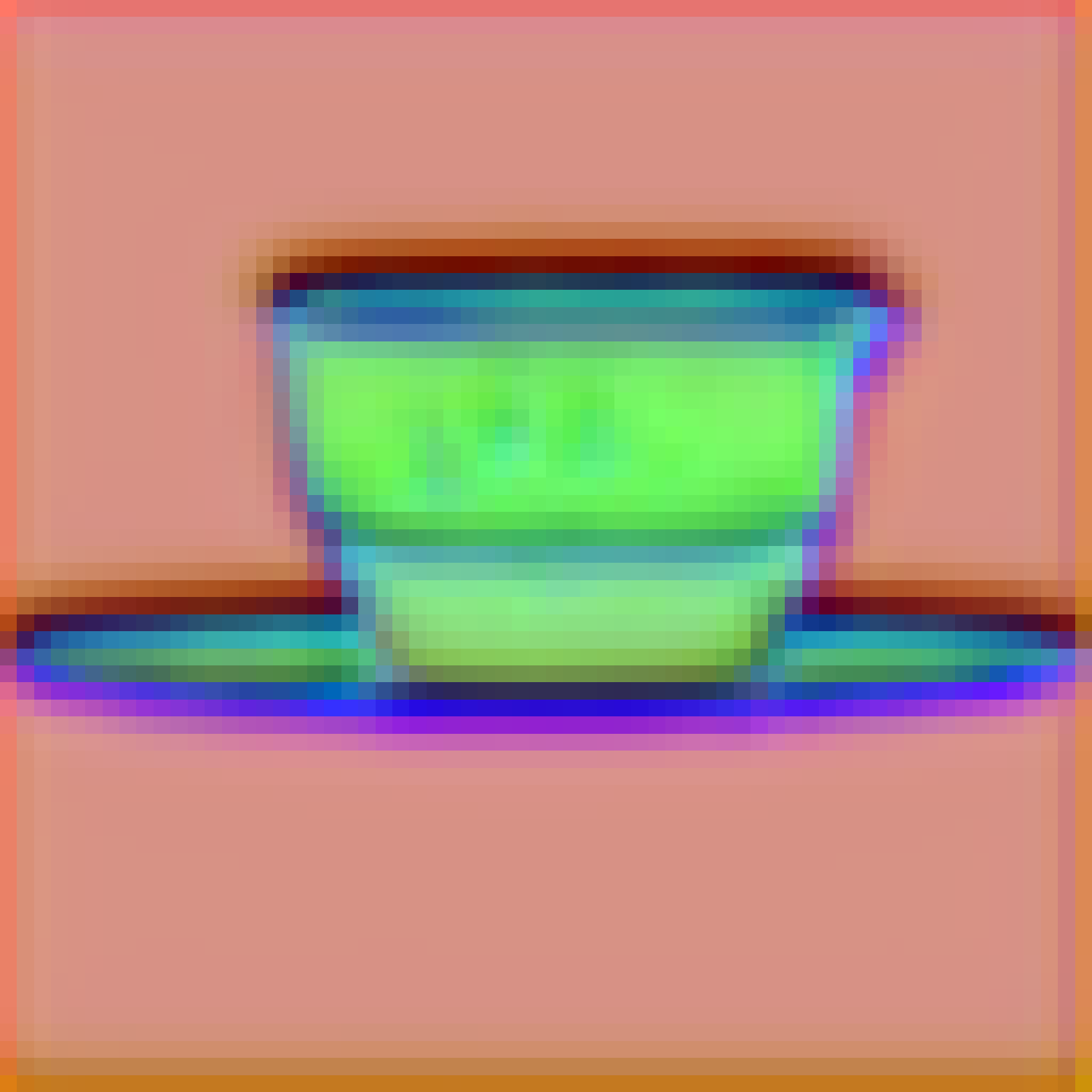} \end{subfigure}%
    \begin{subfigure}[b]{0.142\textwidth} \centering \includegraphics[width=\textwidth]{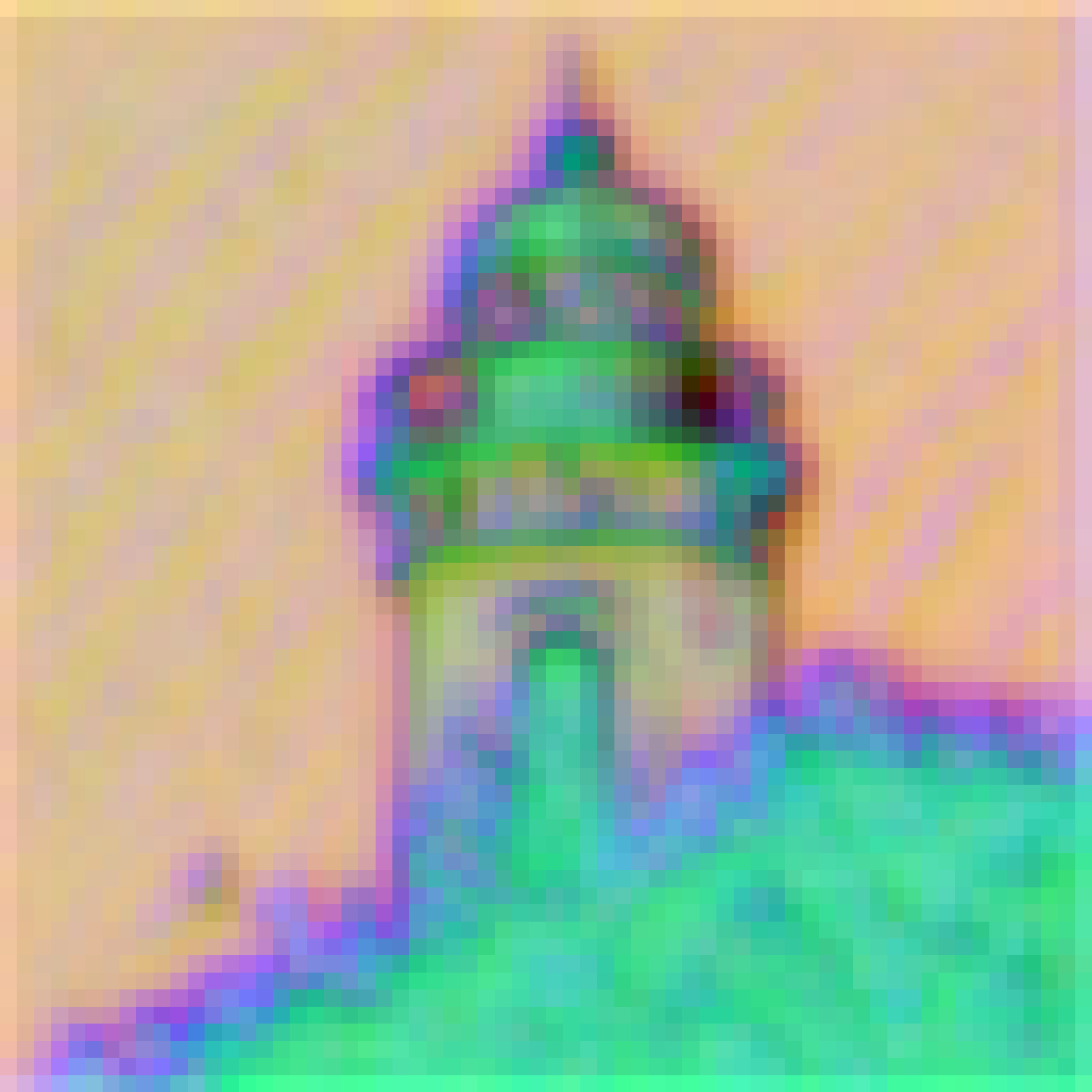} \end{subfigure}%
    \begin{subfigure}[b]{0.142\textwidth} \centering \includegraphics[width=\textwidth]{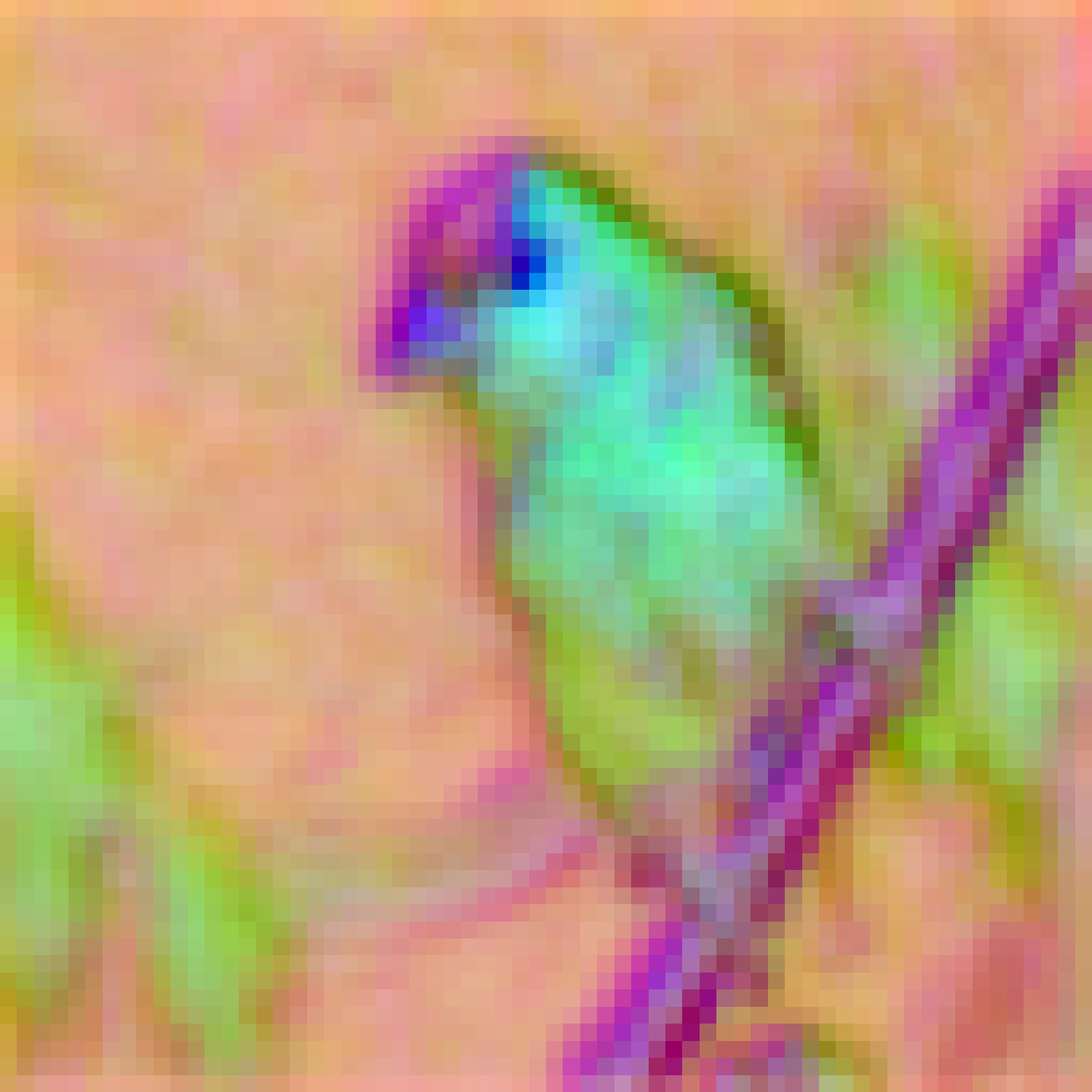} \end{subfigure} \\
    \vspace{-1pt} 

    \begin{subfigure}[b]{0.142\textwidth} \centering \includegraphics[width=\textwidth]{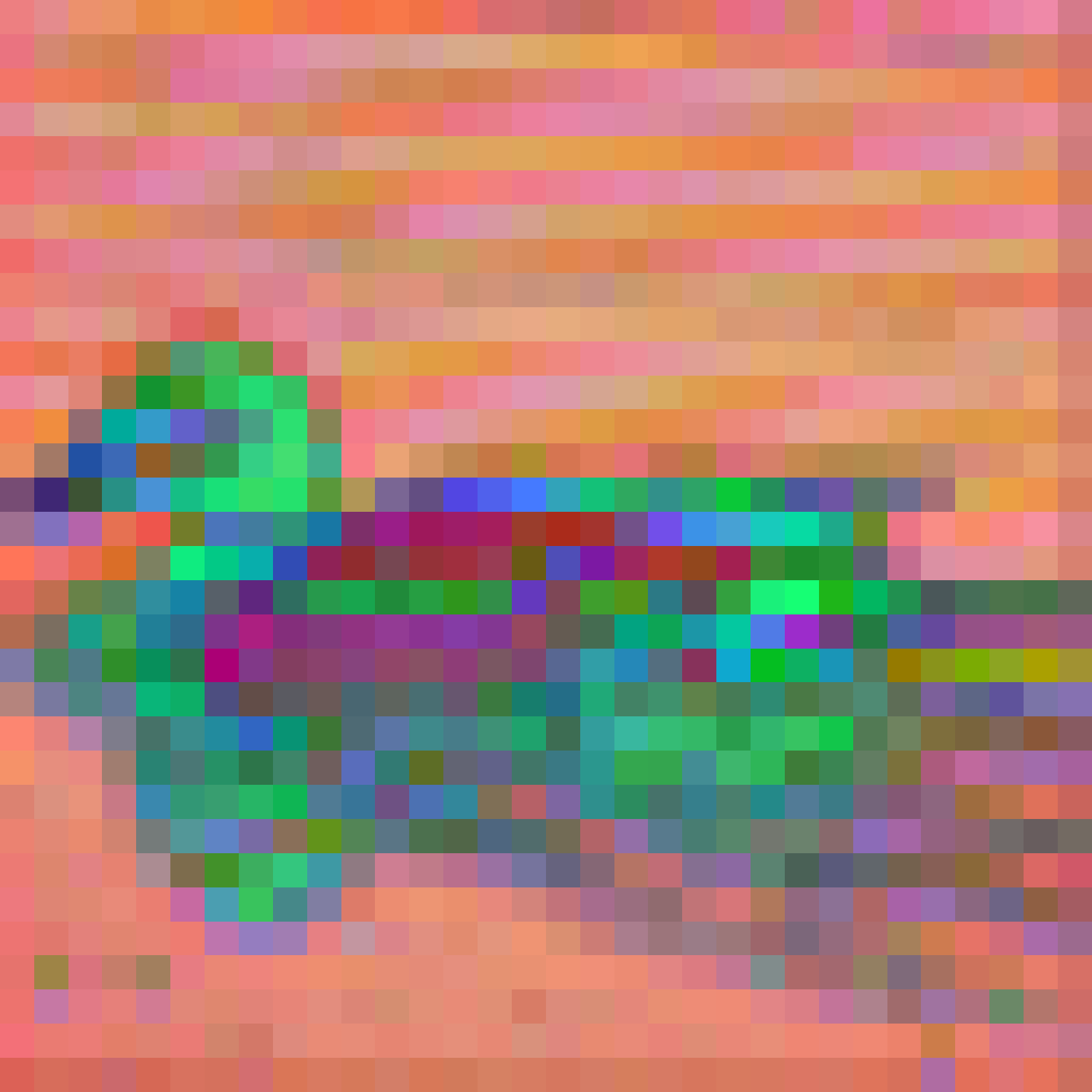} \end{subfigure}%
    \begin{subfigure}[b]{0.142\textwidth} \centering \includegraphics[width=\textwidth]{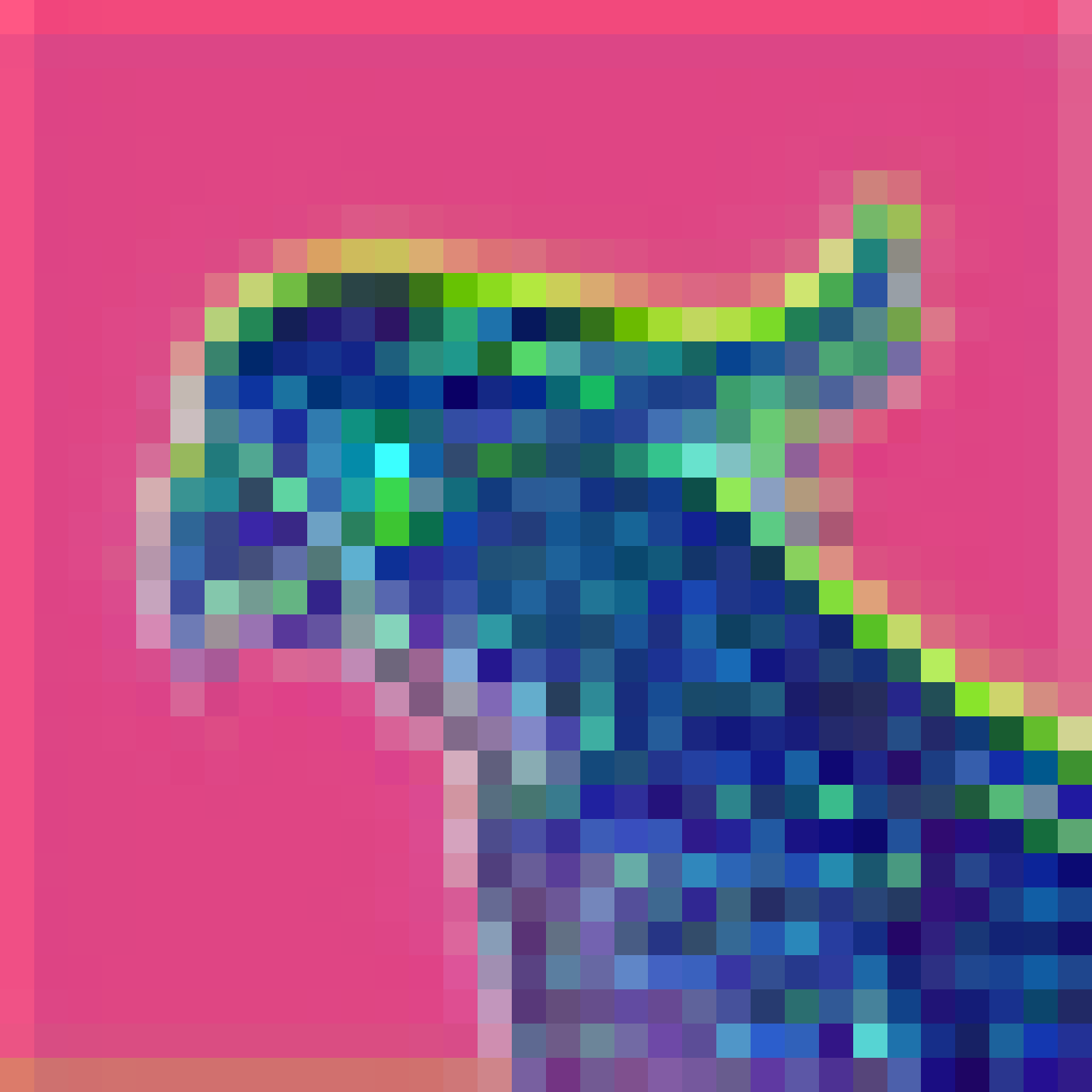} \end{subfigure}%
    \begin{subfigure}[b]{0.142\textwidth} \centering \includegraphics[width=\textwidth]{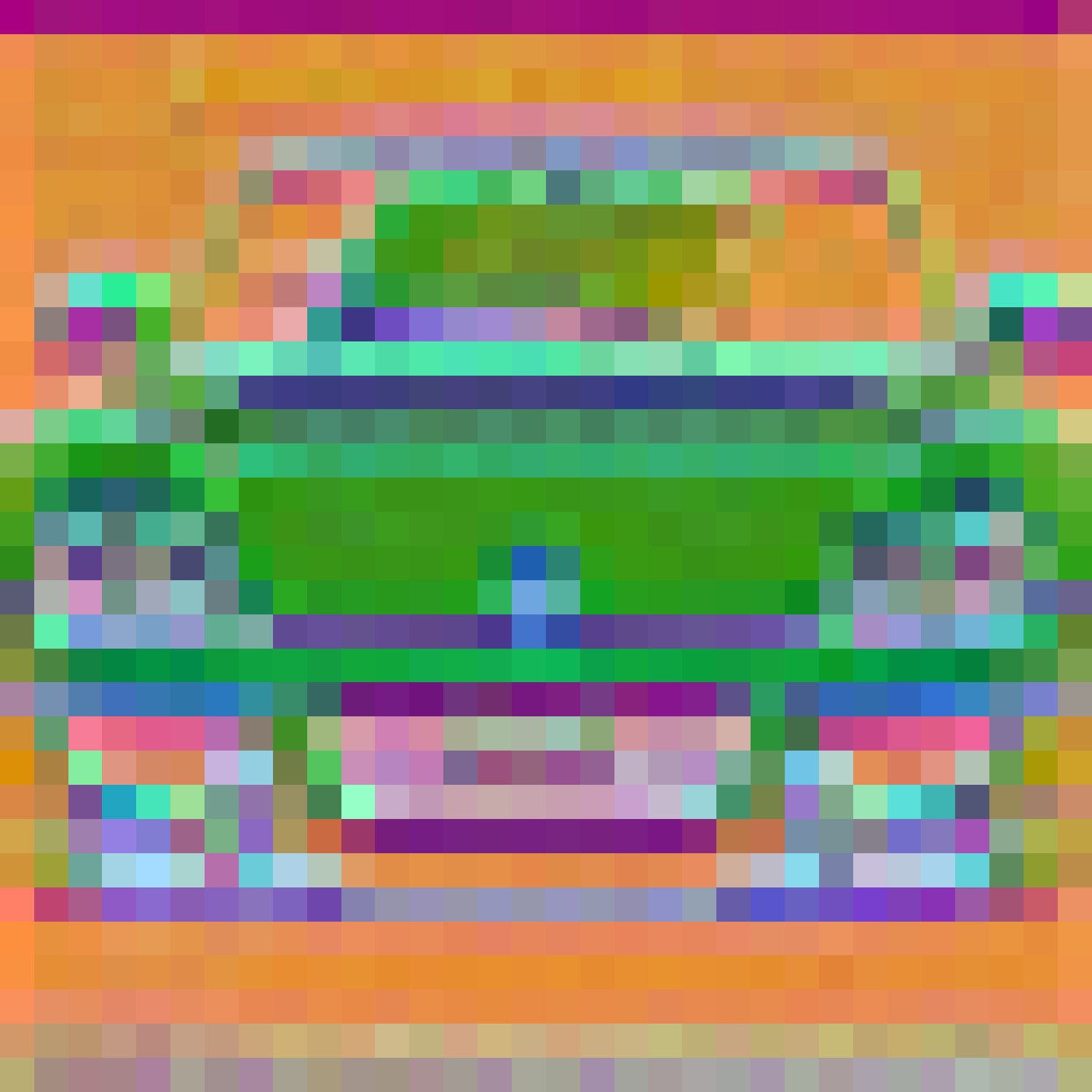} \end{subfigure}%
    \begin{subfigure}[b]{0.142\textwidth} \centering \includegraphics[width=\textwidth]{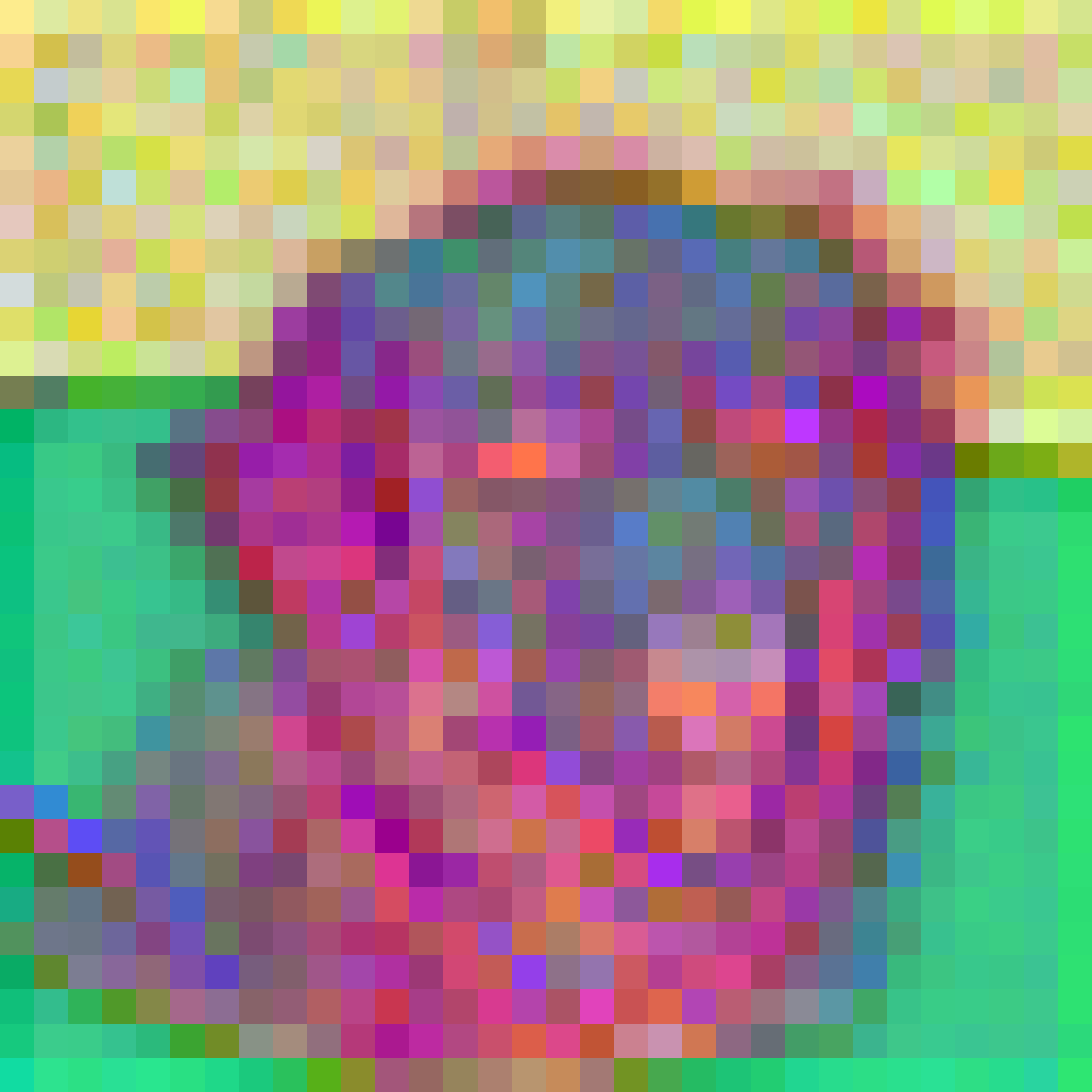} \end{subfigure}%
    \begin{subfigure}[b]{0.142\textwidth} \centering \includegraphics[width=\textwidth]{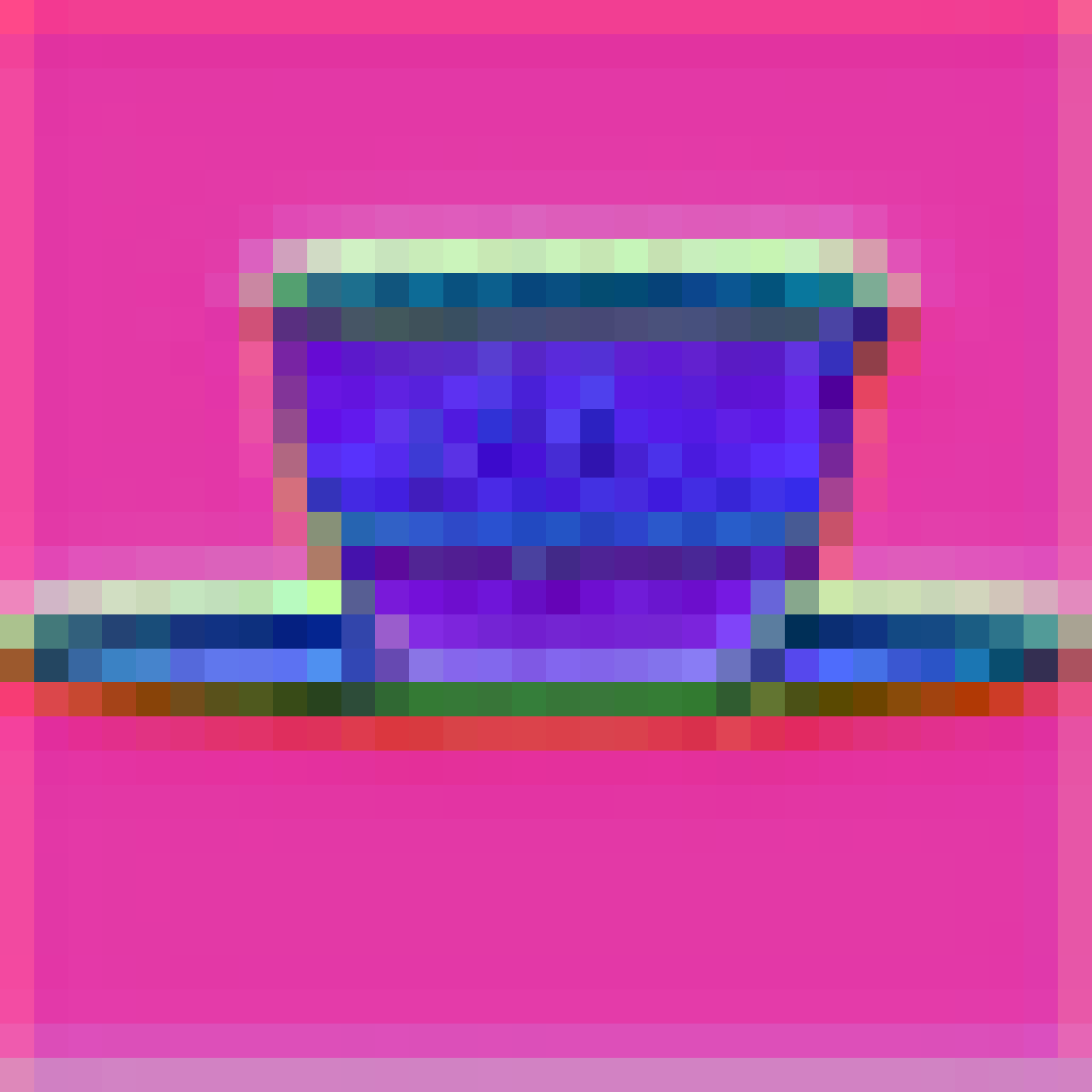} \end{subfigure}%
    \begin{subfigure}[b]{0.142\textwidth} \centering \includegraphics[width=\textwidth]{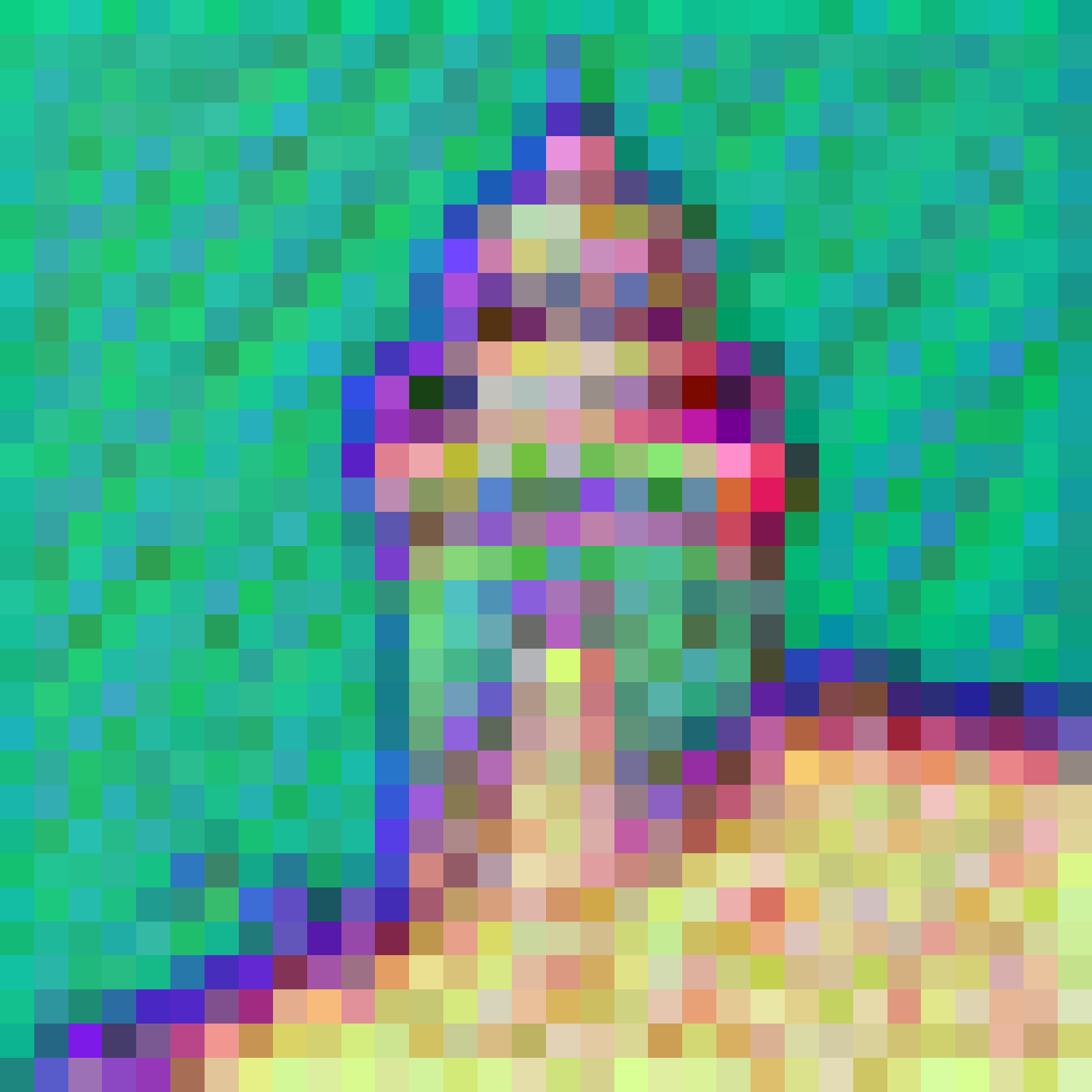} \end{subfigure}%
    \begin{subfigure}[b]{0.142\textwidth} \centering \includegraphics[width=\textwidth]{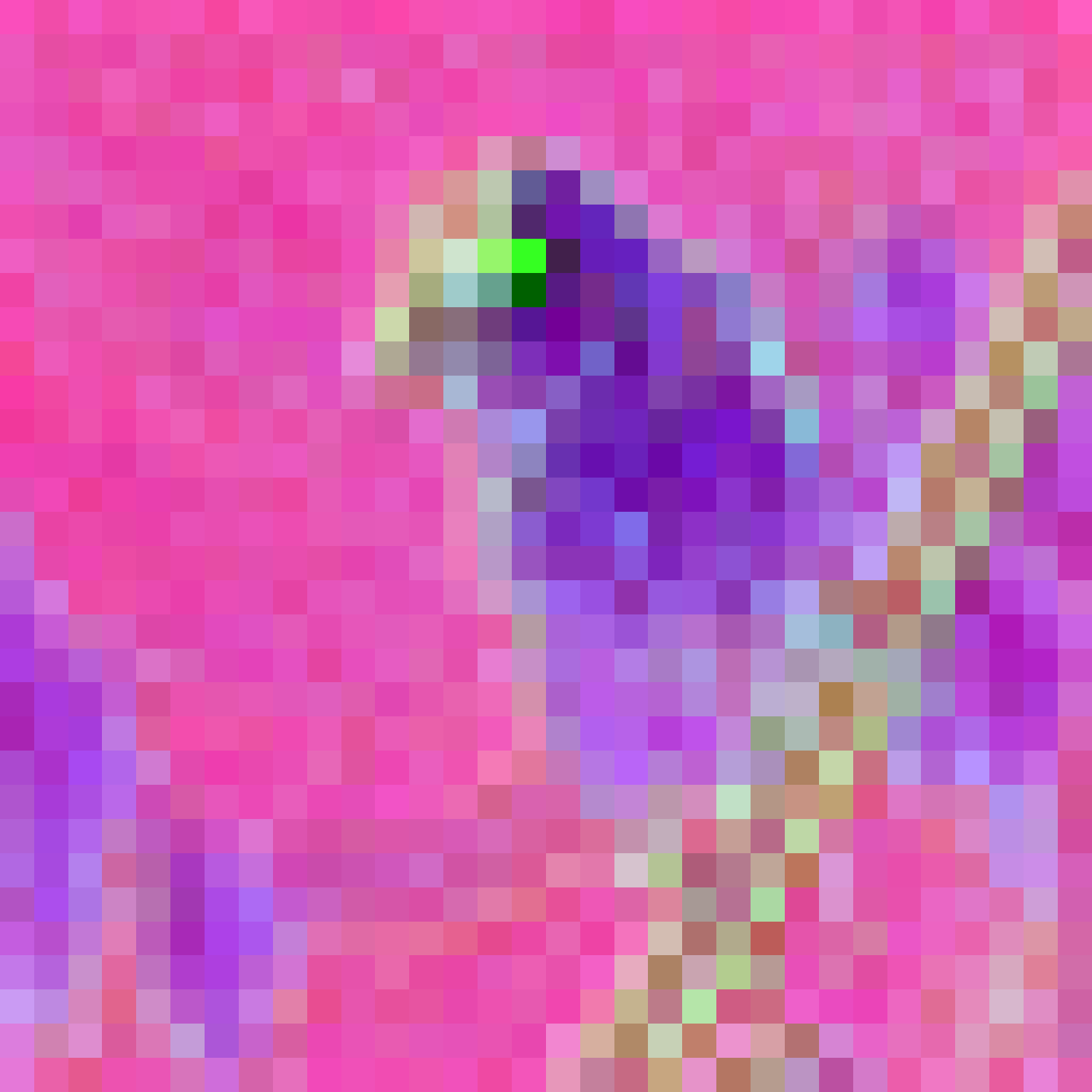} \end{subfigure}

    \caption{\textbf{Group-averaged Feature Visualization via PCA.} From top to bottom: Original images, Middle-level blueprints ($\mathcal{G}_{mid}$), and Deep-level blueprints ($\mathcal{G}_{deep}$). The visualization demonstrates the consistent structural-to-semantic abstraction across diverse categories.}
    \label{fig:vae_pca_transition}
\vspace{-10pt}
\end{figure*}

We leverage the frozen VAE encoder $\mathcal{E}$ as a hierarchical feature extractor to provide the diffusion backbone with a set of multi-scale ``blueprints.'' We partition the VAE encoder layers into four groups ($G_1 \dots G_4$) based on their channel dimensions and spatial resolutions (detailed in App. \ref{app:implementation_details}). While earlier stages ($G_1, G_2$) primarily encapsulate low-level pixel intensities, they exhibit lower alignment purity in our diagnostics (as shown in Sec. \ref{sec:motivation}), likely due to the prevalence of stochastic noise in early denoising phases. In contrast, the selected groups $G_3$ and $G_4$ provide more stable and semantically meaningful guidance for the transformer backbone, further evidenced in Sec.\ref{sec:identify source}. As visualized in Figure~\ref{fig:vae_pca_transition}, the selected VAE stages exhibit a distinct functional stratification across diverse samples:

\textbf{Middle-level Group ($\mathcal{G}_{mid}$):} Derived from $G_3$, this group represents a transitional stage of abstraction, primarily capturing spatial topologies and part-level relationships that provide a structural scaffold for the denoising process.

\textbf{Deep-level Group ($\mathcal{G}_{deep}$):} Derived from $G_4$, this group resides at the peak of the encoder hierarchy, encapsulating broader compositions and global semantic abstractions.

Let $\mathbf{f}_{g, i}$ denote the feature map of the $i$-th layer within group $g \in \{mid, deep\}$. To resolve the spatial discrepancy between VAE stages and the Transformer's patch grid, we apply an adaptive pooling operator $\mathcal{P}(\cdot)$ to unify all features to a fixed latent grid $S \times S$ (e.g., $16 \times 16$):
$\hat{\mathbf{f}}_{g, i} = \mathcal{P}(\mathbf{f}_{g, i}) \in \mathbb{R}^{C_i \times S \times S}$. 
By flattening $\hat{\mathbf{f}}_{g, i}$ into a sequence of length $L=S^2$, each pixel is precisely aligned with the corresponding Transformer token.
\subsection{Collaborative Dynamic Routing}

\textbf{Design Choice: Timestep Conditioning.} We introduce a Dynamic Router $\mathcal{R}_\phi$. While content-adaptive gating is intuitive, we prioritize the timestep $t \in [0, 1]$ as the primary control variable. This is informed by our finding that the shift in demand for hierarchical priors is a trajectory-wide phenomenon dictated by global SNR evolution.

\textbf{Adaptive Weighting and Aggregation.} We employ a linear projection $\text{Embed}(t)$ to predict intra-group aggregation logits $\hat{\boldsymbol{\alpha}}(t)$ and inter-group coordination logits $\hat{\boldsymbol{\beta}}(t)$:
\begin{equation}
\left\{ \hat{\boldsymbol{\alpha}}_{mid}(t), \hat{\boldsymbol{\alpha}}_{deep}(t), \hat{\boldsymbol{\beta}}(t) \right\} = \mathcal{R}_\phi(\text{Embed}(t))
\end{equation}
The aggregated feature $\mathbf{v}_g(t)$ for each group is computed via softmax-normalized summation:
\begin{equation}
\mathbf{v}_g(t) = \sum_{i=1}^{K_g} \frac{\exp(\hat{\alpha}_{g, i}(t))}{\sum_{j=1}^{K_g} \exp(\hat{\alpha}_{g, j}(t))} \cdot \hat{\mathbf{f}}_{g, i}
\end{equation}
To balance middle vs. deep-level contributions, we obtain scaling factors $\boldsymbol{\beta}(t) = \text{Sigmoid}(\hat{\boldsymbol{\beta}}(t))$. The final alignment target $\mathbf{z}_s(t)$ is constructed as:
\begin{equation}
\mathbf{z}_s(t) = \text{Concat}\left[ \beta_{mid}(t) \cdot \mathbf{v}_{mid}(t), \,\, \beta_{deep}(t) \cdot \mathbf{v}_{deep}(t) \right]
\end{equation}
Concatenation preserves representational integrity in distinct channel subspaces, allowing the backbone to adaptively resolve inter-group interactions through the subsequent projection layer.

\subsection{Joint Optimization Objective}
The Transformer $\theta$ and router $\phi$ are optimized jointly. We align the hidden states $h^{(l)}$ from an early block with $\mathbf{z}_s(t)$ using a convolutional projection head $\text{Proj}(\cdot)$. We employ a Cosine Similarity Loss to ensure invariance to activation scales:
\begin{equation}
\mathcal{L}_{align}(\theta, \phi) = \mathbb{E}_{t, x} \left[ 1 - \frac{\text{Proj}(h^{(l)}; \theta) \cdot \mathbf{z}_s(t; \phi)}{\left\| \text{Proj}(h^{(l)}; \theta) \right\|_2 \cdot \left\| \mathbf{z}_s(t; \phi) \right\|_2} \right]
\end{equation}
The total objective is $\mathcal{L}_{total} = \mathcal{L}_{diff} + \lambda \mathcal{L}_{align}$. At inference, the alignment branch is discarded, yielding performance gains with \textbf{zero extra cost}.
\section{Experiments and Analysis}
\label{Sec:Experiments and Analysis}
\subsection{Experimental Setup}
We evaluate AHPA across SiT-B/2, L/2, and XL/2 scales. Following the hierarchical partitioning established in Sec.~\ref{Mining}, the dynamic router $\mathcal{R}_\phi$ is implemented as a 4-layer MLP that adaptively modulates the contributions of $\mathcal{G}_{mid}$ and $\mathcal{G}_{deep}$. The projection head $\text{Proj}(\cdot)$ consists of a single $3\times3$ convolutional layer to preserve spatial inductive biases. All models are trained for 400k iterations on the ImageNet $256\times256$ dataset using NVIDIA A100 GPUs. Detailed architectural configurations, hyperparameter schedules, and evaluation protocols are documented in Appendix~\ref{app:implementation_details}.
\subsection{Ablation Studies}
Experiments in this section are conducted on SiT-B/2. Extended ablations on routing strategies, positional optimality, and component-wise configurations are detailed in Appendix \ref{app:additional_ablations}.

\subsubsection{Determining the Optimal Alignment Stage}

\begin{wraptable}{r}{0.48\textwidth}
    \centering
    \vspace{-1.2em}
    \caption{\textbf{Ablation on alignment block $l$.}}
    \vspace{-8pt}
    \label{tab:layer_ablation}
    \small
    \setlength{\tabcolsep}{3.5pt}
    \begin{tabular}{l@{\hspace{0.5em}}cccc|c}
        \toprule
        $l$ & 2 & \textbf{3} & 4 & 8 & Multi \\
        \midrule
        FID $\downarrow$ & 31.4 & \textbf{30.7} & 31.6 & 32.2 & 32.9 \\
        IS $\uparrow$ & 46.5 & \textbf{48.7} & 47.9 & 44.2 & 47.2 \\
        \bottomrule
    \end{tabular}
    \vspace{-1.2em}
\end{wraptable}

We evaluate the SiT-B/2 backbone's receptivity across different stages using G4 as guidance source. As shown in Table~\ref{tab:layer_ablation}, performance follows a U-shaped trend, with block $l=3$ emerging as the optimal entry point. We observe that early-stage alignment ($l=2$) lacks sufficient abstraction, while late-stage integration ($l \ge 8$) interferes with the model's autonomous denoising. Notably, multi-stage guidance (simultaneously at blocks \{3, 4, 8\}) yields inferior results (32.9 FID), suggesting that redundant constraints introduce optimization conflicts. Consequently, we fix alignment at the 25th percentile of model depth to consistently target the most flexible representational phase.
\subsubsection{Identifying the Optimal Guidance Source}
\label{sec:identify source}
\begin{wraptable}{r}{0.55\textwidth} 
    \centering
    \vspace{-1.2em}
    \caption{\textbf{VAE groups ablation} (Static, $l=3$).}
    \label{tab:group_ablation}
    \small 
    \setlength{\tabcolsep}{6pt} 
    \begin{tabular}{lccc}
        \toprule
        Source & IS $\uparrow$ & FID $\downarrow$ & sFID $\downarrow$ \\
        \midrule
        Vanilla SiT-B/2 & 43.7 & 33.0 & 6.46 \\
        Group 1 & 41.1 & 34.5 & 6.69 \\
        Group 2 & 41.8 & 35.1 & 6.38 \\
        Group 3 ($\mathcal{G}_{mid}$) & 48.9 & 32.6 & 6.45 \\
        Group 4 ($\mathcal{G}_{deep}$) & 48.7 & 30.7 & 6.26 \\
        \rowcolor{gray!10} \textbf{G3 $\cup$ G4}& \textbf{51.6} & \textbf{29.7} & \textbf{6.04} \\
        \bottomrule
    \end{tabular}
    \vspace{-1em}
\end{wraptable}
We investigate the effectiveness of VAE hierarchical groups at $l=3$. Table~\ref{tab:group_ablation} reveals a correlation between semantic depth and generative quality. Aligning with low-level groups (G1, G2) introduces high-frequency noise that distracts optimization. Conversely, mid-to-deep groups (G3, G4) act as effective blueprints, where their combination yields a significant synergy in both structural consistency and semantic accuracy. Based on these findings, we utilize G3 and G4 as our primary alignment sources for AHPA.
\subsection{System-level Comparisons}
\subsubsection{Performance and Efficiency Analysis}

We evaluate AHPA's scalability across multiple model scales (Table~\ref{tab:main_results}) and its competitive standing among current state-of-the-art methods (Table~\ref{tab:sota_compare}). Experiments on ImageNet $512\times512$ and additional qualitative comparisons are detailed in Appendix~\ref{512comparison} and \ref{app:qualitative_comparison}. Our analysis reveals three core insights:

\textbf{Preserving Acceleration Paradigm.} Consistent with the alignment paradigm pioneered by REPA~\cite{yu2024repa}, AHPA successfully recovers the nearly \textbf{17$\times$ } training reduction. By matching the 7M-iteration vanilla baseline at only 400K steps (8.1 vs. 8.3 FID), our results demonstrate that intrinsic VAE hierarchies provide sufficient guidance to maintain the high convergence speeds characteristic of representation alignment, even in the absence of external teachers.

\textbf{Superiority of Adaptive Routing.} AHPA outperforms the latent-based baseline SRA 2 across all scales, with the performance gap widening as model capacity increases. This validates our core hypothesis: while static alignment provides a coarse prior, it fails to accommodate the non-stationary representational needs of larger transformers. By adaptively scheduling hierarchical priors, AHPA resolves the representational mismatch and information collapse that bottleneck previous methods.

\textbf{Competitiveness against Heavyweight Teachers.} AHPA demonstrates strong competitiveness against methods supervised by massive external encoders. Specifically, at 400K iterations, it outperforms REPA in image realism (8.1 vs. 8.5 FID) and achieves results closely approaching REG at 800 epochs (1.40 vs. 1.36 FID). Notably, AHPA maintains superior structural fidelity (4.12 vs. 4.70 sFID against REPA). These results, achieved with negligible overhead (+5\% GFLOPs), prove that adaptively routed intrinsic priors can effectively substitute for external vision learners, eliminating the trade-off between efficiency and synthesis quality.
\begin{table*}[t]
    \centering
    \caption{\textbf{Comprehensive Performance Comparison on ImageNet $256 \times 256$}. We report a holistic evaluation across three model scales (Base, Large, and Extra Large). \textbf{w/o Ext.} indicates whether the method is standalone and does \textit{not} rely on external pre-trained encoders.  \textbf{Iter.} denotes the number of training iterations. Bold indicates the best performance within each model scale.}
    \label{tab:main_results}
    
    \small
    \setlength{\tabcolsep}{0pt} 
    \begin{tabularx}{\textwidth}{@{\extracolsep{\fill}} l l c c c c c c c @{}}
        \toprule
        Scale & Method & w/o Ext. & Iter. & FID $\downarrow$ & sFID $\downarrow$ & IS $\uparrow$ & Prec. $\uparrow$ & Rec. $\uparrow$ \\
        \midrule
        
        \multirow{5}{*}{SiT-B/2} 
        & Vanilla SiT & \greenCheck & 400K & 33.0 & 6.46 & 43.7 & 0.53 & 0.63 \\
        & SRA 2 \cite{wang2026sra2variationalautoencoder} & \greenCheck & 400K & 31.3 & 6.42 & 47.0 & 0.55 & 0.63 \\
        & REPA \cite{yu2024repa} (DINOv2-B) & \redCross & 400K & 27.2 & 6.80 & \textbf{54.9} & 0.57 & \textbf{0.64} \\ \cmidrule{2-9}
        \rowcolor{highlightpink} \cellcolor{white} & \textbf{AHPA (Ours)} & \greenCheck & 400K & \textbf{25.5} & \textbf{5.78} & 52.3 & \textbf{0.59} & 0.63 \\
        
        \midrule
        \multirow{5}{*}{SiT-L/2} 
        & Vanilla SiT & \greenCheck & 400K & 18.8 & 5.29 & 72.0 & 0.64 & \textbf{0.64}  \\
        & SRA 2 \cite{wang2026sra2variationalautoencoder} & \greenCheck & 400K & 16.2 & 5.11 & 77.4 & 0.65 & 0.62 \\
        & REPA \cite{yu2024repa} (DINOv2-B) & \redCross & 400K & 12.6 & 5.25 & \textbf{95.0} & 0.67 & \textbf{0.64} \\ \cmidrule{2-9}
        \rowcolor{highlightpink} \cellcolor{white} & \textbf{AHPA (Ours)} & \greenCheck & 400K & \textbf{10.5} & \textbf{4.97} & 92.1 & \textbf{0.69} & 0.62 \\
        
        \midrule
        \multirow{5}{*}{SiT-XL/2} 
        & Vanilla SiT & \greenCheck & 400K & 17.2 & 5.95 & 83.1 & 0.64 & 0.63 \\
        & Vanilla SiT & \greenCheck & 7M & 8.3 & 6.32 & \textbf{131.7} & 0.68 & \textbf{0.67} \\
        & SRA 2 \cite{wang2026sra2variationalautoencoder} & \greenCheck & 400K & 13.8 & 5.10 & 94.3 & 0.65 & 0.64 \\
        & REPA \cite{yu2024repa} (DINOv2-B) & \redCross & 400K & 8.5 & 5.14 & 110.6 & 0.69 & 0.64 \\ \cmidrule{2-9}
        \rowcolor{highlightpink} \cellcolor{white} & \textbf{AHPA (Ours)} & \greenCheck & 400K & \textbf{8.1} & \textbf{4.90} & 112.0 & \textbf{0.71} & 0.62 \\
        \bottomrule
    \end{tabularx}
\end{table*}
\begin{table*}[t]
\centering
\caption{\textbf{State-of-the-Art Comparison on ImageNet $256\times256$ with CFG.} Bold indicates the best results among methods without external dependencies.}
\vspace{-8pt}
\label{tab:sota_compare}
\small 
\setlength{\tabcolsep}{2pt} 
\begin{tabularx}{\textwidth}{@{\extracolsep{\fill}} lccccccc @{}}
\toprule
Method & Epochs & FID$\downarrow$ & sFID$\downarrow$ & IS$\uparrow$ & Pre.$\uparrow$ & Rec.$\uparrow$ & w/o Ext. \\ \midrule

\textit{Latent diffusion, Transformer} & & & & & & & \\
DiT-XL/2 \cite{peebles2023scalable} & 1400 & 2.27 & 4.60 & 278.2 & \textbf{0.83} & 0.57 & \greenCheck \\
+ iSSD \cite{Ma2025iSSD} & 1400+0.8 & 2.02 & 4.22 & 250.0 & 0.81 & 0.60 & \redCross \\
MaskDiT \cite{zheng2024maskdit} & 1600 & 2.28 & 5.67 & 276.6 & 0.80 & 0.61 & \redCross \\
SD-DiT \cite{zhu2024SD-DiT} & 480 & 3.23 & - & - & - & - & \redCross \\ \addlinespace
SiT-XL/2 \cite{ma2024sit} & 1400 & 2.06 & 4.50 & 270.3 & 0.82 & 0.59 & \greenCheck \\
+ REPA \cite{yu2024repa} & 800 & 1.42 & 4.70 & 311.4 & 0.80 & 0.63 & \redCross \\
+ REG \cite{wu2025reg} & 800 & 1.36 & 4.25 & 299.4 & 0.77 & 0.66 & \redCross \\
+ SRA \cite{jiang2026sra} & 800 & 1.58 & 4.65 & 305.7 & 0.80 & 0.63 & \greenCheck \\
+ SRA 2 \cite{wang2026sra2variationalautoencoder} & 800 & 1.52 & 4.63 & 316.2 & 0.82 & 0.62 & \greenCheck \\ \addlinespace
\rowcolor{highlightpink} \cellcolor{white}+ \textbf{AHPA (Ours)} & 100 & 2.15 & 4.38 & 265.4 & 0.81 & 0.58 & \greenCheck \\
\rowcolor{highlightpink} \cellcolor{white}+ \textbf{AHPA (Ours)} & 200 & 1.78 & 4.26 & 291.8 & 0.82 & 0.61 & \greenCheck \\
\rowcolor{highlightpink} \cellcolor{white}+ \textbf{AHPA (Ours)} & 400 & 1.56 & 4.15 & 308.2 & 0.81 & 0.63 & \greenCheck \\
\rowcolor{highlightpink} \cellcolor{white}+ \textbf{AHPA (Ours)} & 800 & \textbf{1.40} & \textbf{4.12} & \textbf{317.5} & 0.81 & \textbf{0.64} & \greenCheck \\ \bottomrule
\end{tabularx}
\vspace{-15pt}
\end{table*}
\subsubsection{Text-to-Image Generation on MS-COCO}
\begin{wraptable}{r}{0.45\textwidth} 
    \centering
    \vspace{-1.2em} 
    \caption{\textbf{T2I on MS-COCO.} Evaluated using MMDiT (SDE, NFE=250).}
    \vspace{-8pt}
    \label{tab:t2i_brief}
    \small
    \setlength{\tabcolsep}{4pt}
    \begin{tabular}{l|cc}
        \toprule
        Method & FID $\downarrow$ & PS $\uparrow$ \\
        \midrule
        MMDiT & 5.30 & 20.54 \\
        + REPA & 4.74 & 20.90 \\
        + SRA 2 & 4.91 & 20.76 \\
        \rowcolor{highlightpink} \textbf{+ AHPA} & \textbf{4.52} & \textbf{21.33} \\ 
        \bottomrule
    \end{tabular}
    \vspace{-1em} 
\end{wraptable}
To verify the generalization of AHPA, we conduct text-to-image (T2I) generation experiments on MS-COCO using the MMDiT backbone. As shown in Table \ref{tab:t2i_brief}, under SDE sampling (NFE=250), AHPA achieves a FID of \textbf{4.52} and a PickScore (PS) of \textbf{21.03}. Despite relying solely on intrinsic VAE features, AHPA not only outperforms previous self-alignment baselines (SRA 2) but also surpasses methods utilizing heavy-weight external vision learners (REPA) in both image realism and text-alignment. These results confirm that adaptively scheduled hierarchical priors provide robust guidance for T2I task. Detailed experimental setups and ODE sampler results are provided in Appendix \ref{app:t2i_exp}.
\subsubsection{Training Computational Efficiency}

\begin{wraptable}{r}{0.5\textwidth}
    \centering
    \vspace{-1.5em}
    \caption{\textbf{Cost comparison on ImageNet $256^2$}. TS: Training Speed (batch size 256, A100 GPUs).}
    \vspace{-5pt}
    \label{tab:cost_comparison}
    \small
    \setlength{\tabcolsep}{3pt}
    \begin{tabular}{l c c c}
        \toprule
        Method & \#EFP (M) & TS $\uparrow$ & GFLOPs $\downarrow$ \\
        \midrule
        Baseline (SiT-XL/2) & $0$ & 8.1 & 114.5 \\
        + REPA \cite{yu2024repa} & $94$ & 6.3 & 138.5 \\
        + SRA \cite{jiang2026sra} & $483$ & 5.1 & 197.6 \\
        + SRA 2 \cite{wang2026sra2variationalautoencoder} & $18$ & 7.2 & 118.6 \\
        \rowcolor{highlightpink} \textbf{+ AHPA (Ours)} & \textbf{2.5} & \textbf{7.4} & \textbf{119.9} \\
        \bottomrule
    \end{tabular}
    \vspace{-1em}
\end{wraptable}

We evaluate the training overhead by pre-computing VAE hierarchical features offline, eliminating real-time encoding latency. As shown in Table~\ref{tab:cost_comparison}, AHPA demonstrates superior efficiency over teacher-based methods. While REPA incurs substantial overhead (up to +73\% GFLOPs) due to heavy external encoders, AHPA is \textit{external-free}. By employing a lightweight MLP router and a single-layer projection head, AHPA adds only a negligible \textbf{5\%} to total GFLOPs and 2.5M parameters. Our method maintains a high throughput of 7.4 Iter/s, nearly on par with the baseline, providing a frugal yet effective path for representation alignment. Detailed comparison is presented in Appendix ~\ref{app:detailed_cost_comparison}.
\subsection{Mechanistic Insights into Non-stationary Alignment}
\label{sec:mechanics}

In this section, we provide empirical evidence for how AHPA resolves the representational mismatch.

\subsubsection{Emergent Routing Policy and Physical Consistency}

We analyze the trajectories learned by the Dynamic Router $\mathcal{R}_\phi$ to uncover the model's emergent alignment preferences. Crucially, the router is optimized solely via the generative objective ($\mathcal{L}_{diff} + \lambda\mathcal{L}_{align}$), with no explicit knowledge of the G-SNR diagnostic conducted in Sec.~\ref{sec:motivation}.

\begin{figure*} 
  \centering
  \includegraphics[width=\linewidth]{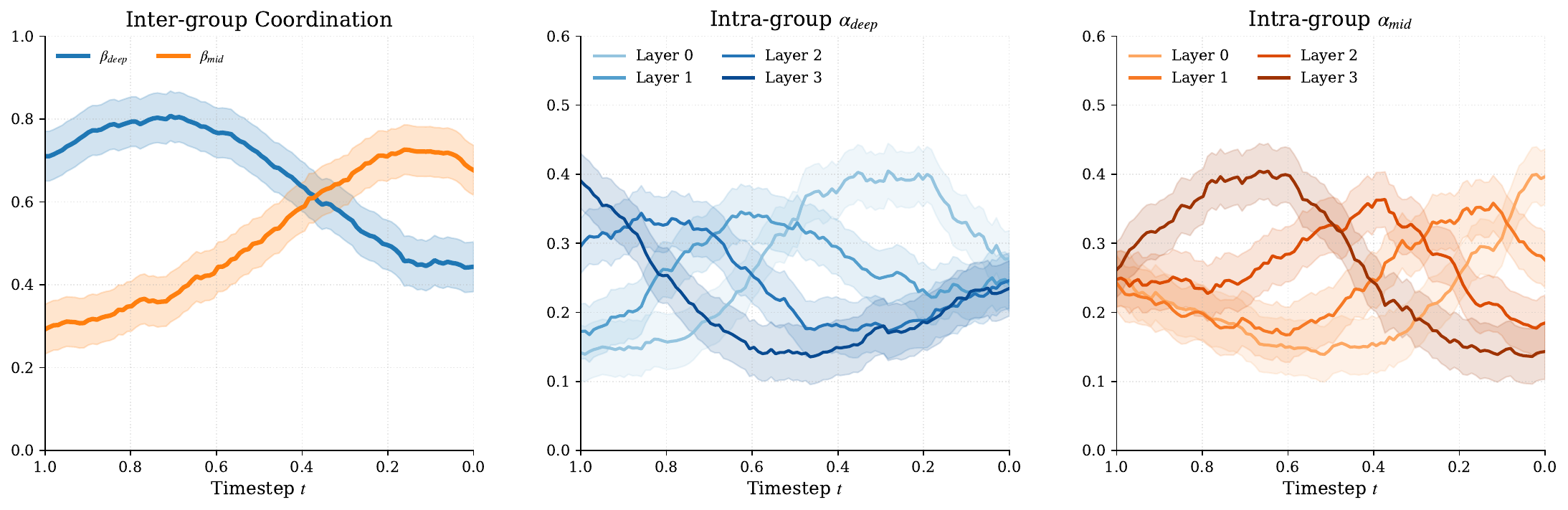} 
  \caption{\textbf{Mechanistic analysis of AHPA's dynamic routing policy.} 
  (Left) The inter-group weights $\beta$ exhibit a clear transition from semantic anchoring to structural refinement as $t \to 0$. 
  (Middle, Right) The intra-group weights $\alpha$ show the microscopic selection of hierarchical layers within each functional group. Results are obtained from SiT-XL/2 after 400k training iterations on ImageNet.}
  \label{fig:routing_analysis}
  \vspace{-10pt}
\end{figure*}

\textbf{Autonomous Handover Policy.} As visualized in Fig.~\ref{fig:routing_analysis} (Left), the macroscopic weights reveal an autonomous handover policy that mirrors the underlying SNR evolution. The router prioritizes $\beta_{deep}$ during high-noise phases to anchor the generative manifold, while spontaneously shifting focus toward $\beta_{mid}$ as denoising progresses. This self-adaptive transition provides a powerful cross-validation: the fact that the backbone ``chooses'' to reconfigure its guidance source to minimize loss confirms that the non-stationary demand aligns with the inherent dynamics of the diffusion process.

\textbf{Microscopic Calibration.} Within each group, the router executes a fine-grained granularity shifting mechanism. As shown in Fig.~\ref{fig:routing_analysis} (Middle, Right), layer-wise focus consistently transitions from deep bottleneck features to shallower ones as $t \to 0$. This synchronized evolution ensures that the hierarchical priors remain closely matched to the transformer's shrinking receptive field and its increasing need for granular structural cues during final synthesis.

\subsubsection{Fidelity Synergy: AHPA Resolves Representational Mismatch}
\label{sec:fidelity_analysis}

\begin{wrapfigure}{r}{0.5\textwidth}
    \centering
    \vspace{-15pt}
    \small
    \captionof{table}{Analysis of alignment fidelity (G-SNR). Avg. denotes the mean of 100 $t$-samples.}
    \label{tab:mismatch_analysis}
    \setlength{\tabcolsep}{4pt}
    \begin{tabular}{lccc}
        \toprule
        Method & $t=0.9$ & $t=0.3$ & Avg. \\
        \midrule
        REPA (DINOv2-B) & 13.9 & 4.2 & 8.5 \\
        SRA 2 (Latent)    & 12.0 & 3.1 & 7.2 \\
        Uniform $\mathcal{G}_{deep}$ & 12.3 & 4.1 & 7.3 \\
        Uniform $\mathcal{G}_{mid}$  & 4.1  & 9.2 & 7.0 \\
        \midrule
        \rowcolor{highlightpink} \textbf{AHPA (Ours)} & 11.9 & \textbf{10.8} & \textbf{11.6} \\
        \bottomrule
    \end{tabular}
    \vspace{-10pt}
\end{wrapfigure}

We analyze AHPA's resolution of representational mismatch through the directional consistency of its alignment gradients (G-SNR). As shown in Table~\ref{tab:mismatch_analysis}, static methods suffer from fidelity collapse in late denoising stages, where fixed priors misalign with the model's evolving receptivity and introduce destructive interference. In contrast, AHPA maintains a healthy guidance spectrum throughout the trajectory, achieving the highest averaged G-SNR (11.6).
By dynamically routing mid-level structural priors as the backbone's focus shifts, AHPA keeps the alignment signal constructive during critical refinement phases, mitigating the gradient impurity that bottlenecks static strategies. The sustained high-fidelity gradients offer a mechanistic explanation for the improved convergence and structural integrity in our main results, confirming that resolving this non-stationary mismatch is key to unlocking the full potential of intrinsic hierarchical guidance.
\section{Conclusion}

In this study, we introduced AHPA, an adaptive hierarchical alignment framework that resolves the representational mismatch inherent in static guidance paradigms. By identifying the non-stationary alignment requirement along the diffusion trajectory, we demonstrated that the model’s needs shift from global semantic anchoring to fine-grained structural refinement. AHPA effectively addresses this by mining hierarchical priors from the intrinsic VAE encoder and adaptively scheduling them via a Dynamic Router. Extensive experiments show that AHPA achieves competitive convergence and structural fidelity, matching or exceeding the performance of heavy-weight external teachers. As a frugal and self-contained solution, AHPA offers a principled path for accelerating large-scale generative models without the burden of external dependencies or redundant computational overhead.


\bibliographystyle{plain} 
\bibliography{main}       

\newpage 
\appendix 

\renewcommand{\thefigure}{\thesection.\arabic{figure}}
\renewcommand{\thetable}{\thesection.\arabic{table}}

\section{Detailed Experimental Settings} \label{app:implementation_details}

\textbf{Model Configurations.} We evaluate AHPA across three scales: SiT-B/2, L/2, and XL/2. The pre-trained VAE encoder is frozen and utilized as a hierarchical feature extractor. Following the observation that deep convolutional networks exhibit a progressive transition from local textures to global abstractions, We partition the 15 VAE encoder layers into four functional groups ($G_1 \dots G_4$) based on their channel dimensions and spatial resolutions. This partitioning is conducted based on the standard SD-VAE architecture \cite{rombach2022LDM-4}, as detailed in Table~\ref{tab:vae_groups}.

This partitioning reflects the inherent representational hierarchy of the encoder: $G_1$ and $G_2$ reside in the early stages, primarily capturing local pixel intensities and geometric primitives. As the network deepens, the representations undergo a gradual abstraction process: $G_3$ aligns with a transitional stage that preserves spatial topologies and structural relationships, while $G_4$ resides at the deepest level, encapsulating broader semantic compositions. By designating $G_3$ and $G_4$ as $\mathcal{G}_{mid}$ and $\mathcal{G}_{deep}$ respectively, our Dynamic Router can adaptively bridge the gap between these varying levels of abstraction and the diffusion process throughout the denoising trajectory.

\begin{table}[h]
    \centering
    \caption{\textbf{VAE Encoder Layer Partitioning.} Layers are grouped by channel dimensions ($C$), native spatial resolutions, and their representational focus based on the SD-VAE architecture. Groups $G_3$ and $G_4$ are utilized as $\mathcal{G}_{mid}$ and $\mathcal{G}_{deep}$ respectively for adaptive guidance.}
    \label{tab:vae_groups}
    \begin{small}
    \begin{tabularx}{\textwidth}{lc c c p{3.5cm}}
        \toprule
        Group ID & VAE Encoder Layers & $C$ & Native Res. & Representational Focus \\
        \midrule
        $G_1$ & \begin{tabular}[c]{@{}l@{}}\texttt{conv\_in, down\_0\_block\_0,}\\ \texttt{down\_0\_block\_1, down\_0\_downsample}\end{tabular} & 128 & $256^2 \to 128^2$ & Local pixel textures \\
        \addlinespace[0.5em]
        $G_2$ & \begin{tabular}[c]{@{}l@{}}\texttt{down\_1\_block\_0, down\_1\_block\_1,}\\ \texttt{down\_1\_downsample}\end{tabular} & 256 & $128^2 \to 64^2$ & Geometric primitives \\
        \addlinespace[0.5em]
        $G_3$ ($\mathcal{G}_{mid}$) & \begin{tabular}[c]{@{}l@{}}\texttt{down\_2\_block\_0, down\_2\_block\_1,}\\ \texttt{down\_2\_downsample, down\_3\_block\_0}\end{tabular} & 512 & $64^2 \to 32^2$ & Structural topologies \\
        \addlinespace[0.5em]
        $G_4$ ($\mathcal{G}_{deep}$) & \begin{tabular}[c]{@{}l@{}}\texttt{down\_3\_block\_1, mid\_block\_1,}\\ \texttt{mid\_attn\_1, mid\_block\_2}\end{tabular} & 512 & $32^2$ & Semantic compositions \\
        \bottomrule
    \end{tabularx}
    \end{small}
\end{table}

\textbf{Alignment Modules.} To align multi-resolution VAE features with the transformer patch grid, all selected features are spatially resampled to $16 \times 16$ via bilinear interpolation. The dynamic router $\mathcal{R}_\phi$ is implemented as a 4-layer MLP that takes the timestep $t$ as input to predict hierarchical weights. For feature projection, we employ a single-layer $3 \times 3$ convolution as the projection head. This choice preserves spatial inductive biases and smooths potential resampling artifacts while ensuring a lightweight enhancement.

\textbf{Training Protocol.} All models are trained for 400k iterations. We utilize the AdamW optimizer with a constant batch size and learning rate as specified in Table~\ref{tab:hyperparams}. To isolate the intrinsic contribution of hierarchical priors, we perform alignment training \textit{without} Classifier-Free Guidance (CFG). All experiments are conducted on NVIDIA A100 GPUs.

\begin{table}[h]
\centering
\caption{\textbf{Hyperparameter settings across different model scales for AHPA.} We detail the architectural configurations, optimization schedules, and alignment parameters used in our experiments.}
\label{tab:hyperparams}
\small
\setlength{\tabcolsep}{8pt}
\begin{tabular}{lccc}
\toprule
\textbf{Configuration} & \textbf{SiT-B/2} & \textbf{SiT-L/2} & \textbf{SiT-XL/2} \\ \midrule
\textit{Architecture} & & & \\
\# Parameters & 132M & 460M & 677M \\
Input Latent & $32\times32\times4$ & $32\times32\times4$ & $32\times32\times4$ \\
Layers & 12 & 24 & 28 \\
Hidden dim. & 768 & 1,024 & 1,152 \\
Num. heads & 12 & 16 & 16 \\ \midrule
\textit{AHPA Settings (Dynamic Router)} & & & \\
Alignment Weight $\lambda$ & 1.0 & 1.0 & 1.0 \\
Alignment Depth & 3 & 6 & 7 \\
Router MLP Layers & 4 & 4 & 4 \\
Router Time Embed Dim. & 128 & 128 & 128 \\
Router Hidden Dim. & 256 & 256 & 256 \\
Projection Head & $3\times3$ Conv & $3\times3$ Conv & $3\times3$ Conv \\
Encoder $\mathcal{E}$ (Frozen) & VAE-Enc & VAE-Enc & VAE-Enc \\ \midrule
\textit{Optimization} & & & \\
Batch size & 256 & 256 & 256 \\
Optimizer & AdamW & AdamW & AdamW \\
Learning rate & $2 \times 10^{-4}$ & $2 \times 10^{-4}$ & $2 \times 10^{-4}$ \\
$(\beta_1, \beta_2)$ & (0.9, 0.999) & (0.9, 0.999) & (0.9, 0.999) \\ 
Weight decay & 0.0 & 0.0 & 0.0 \\ 
Max Grad Norm & 2.0 & 2.0 & 2.0 \\
EMA decay & 0.9999 & 0.9999 & 0.9999 \\ 
CFG Dropout Prob. (Training) & 0.1 & 0.1 & 0.1 \\ \midrule
\textit{Interpolants \& Sampling} & & & \\
$\alpha_t$ & $1 - t$ & $1 - t$ & $1 - t$ \\
$\sigma_t$ & $t$ & $t$ & $t$ \\
Training objective & $v$-prediction & $v$-prediction & $v$-prediction \\
Sampler & Euler & Euler & Euler \\
Sampling steps (NFE) & 250 & 250 & 250 \\  \bottomrule
\end{tabular}
\end{table}

\textbf{Evaluation Protocol.} To assess generative quality, we employ FID for realism, sFID for spatial coherence, and IS for diversity, alongside Precision and Recall. All metrics are computed using 50,000 samples against the ImageNet training set. During inference, we use the deterministic ODE solver with 250 steps to ensure consistency across all scales and baselines.
\section{Generalization Across VAE Architectures} 

To verify whether the learned routing policy $\phi$ is over-fitted to the specific characteristics of SD-VAE, we conduct a cross-architecture study using SDXL-VAE \cite{podell2023SDXL-VAE} and Consistency VAE \cite{kim2024consistencyvae}. We adopt a resolution-centric grouping strategy: layers with the lowest spatial resolution (e.g., $32 \times 32$ for a $256^2$ input) are assigned to $\mathcal{G}_{deep}$, while layers from the preceding stage (e.g., $64 \times 64$) form $\mathcal{G}_{mid}$. 

\textbf{Zero-shot Alignment Health.} We first evaluate the policy's ability to coordinate "unseen" feature distributions. As shown in Table~\ref{tab:vae_generalization}, we evaluate the AHPA-trained router on a pre-trained vanilla SiT-B/2 by swapping the supervisor VAE encoder without any further fine-tuning. AHPA maintains a robust average G-SNR (exceeding 10.0), significantly outperforming the static uniform baselines. This high zero-shot alignment health confirms that AHPA captures a universal representational gradient inherent in convolutional hierarchies, proving that our dynamic scheduling logic is an intrinsic property of the diffusion trajectory rather than an artifact of a specific encoder.

\textbf{Generative Performance Gain.} To further assess whether this alignment health translates into superior image quality, we train SiT-B/2 from scratch for 400k iterations with each VAE type. As summarized in Table~\ref{tab:vae_fid_results}, AHPA consistently yields significant improvements in both FID and sFID across all architectures. Notably, even with the high-performance Consistency VAE, AHPA achieves a notable FID reduction compared to its vanilla counterpart. These results demonstrate that AHPA effectively mitigates representational bottlenecks regardless of the underlying latent manifold, showcasing its robustness as a general-purpose guidance framework.

\begin{table}[h]
\centering
\caption{\textbf{Zero-shot Alignment Diagnostics.} All metrics are Trajectory Averages (G-SNR) computed over 100 $t$-samples on a frozen vanilla SiT-B/2 backbone. AHPA maintains high alignment health despite supervisor swapping.}
\label{tab:vae_generalization}
\small
\begin{tabularx}{\columnwidth}{@{\extracolsep{\fill}} l l c @{}}
\toprule
Backbone (Frozen) & Supervisor VAE Encoder & Avg. G-SNR $\uparrow$ \\
\midrule
SiT-B/2 (Vanilla) & --- & --- \\
\midrule
\rowcolor{gray!20} \cellcolor{white} & SD-VAE \cite{rombach2022SDVAE} (Default) & \textbf{11.6} \\
\rowcolor{gray!20} \cellcolor{white} & SDXL-VAE \cite{podell2023SDXL-VAE} & 10.4 \\
\rowcolor{gray!20} \cellcolor{white} \multirow{-3}{*}{\textbf{AHPA (Ours)}} & Consistency VAE \cite{kim2024consistencyvae} & 10.1 \\
\midrule
Uniform $\mathcal{G}_{deep}$ & SD-VAE & 7.3 \\
Uniform $\mathcal{G}_{mid}$ & SD-VAE & 7.0 \\
\bottomrule
\end{tabularx}
\end{table}

\begin{table}[h]
\centering
\caption{\textbf{Performance Comparison across VAE Architectures.} All models are SiT-B/2 trained from scratch for 400k iterations on ImageNet $256 \times 256$. AHPA consistently enhances generative quality across different latent manifolds.}
\label{tab:vae_fid_results}
\small
\begin{tabular}{l l c c}
\toprule
Supervisor VAE & Method & FID $\downarrow$ & sFID $\downarrow$ \\
\midrule
 & Vanilla SiT & 33.0 & 6.46 \\
\rowcolor{gray!20}
\cellcolor{white} \multirow{-2}{*}{SD-VAE (Default)} & \textbf{AHPA (Ours)} & \textbf{25.5} & \textbf{5.78} \\
\midrule
 & Vanilla SiT & 28.4 & 6.12 \\
\rowcolor{gray!20}
\cellcolor{white} \multirow{-2}{*}{SDXL-VAE} & \textbf{AHPA (Ours)} & \textbf{22.1} & \textbf{5.45} \\
\midrule
 & Vanilla SiT & 24.5 & 5.90 \\
\rowcolor{gray!20}
\cellcolor{white} \multirow{-2}{*}{Consistency VAE} & \textbf{AHPA (Ours)} & \textbf{21.7} & \textbf{5.48} \\
\bottomrule
\end{tabular}
\end{table}
\section{Text-to-Image Generation Experiment} \label{app:t2i_exp}

To verify the generalization of AHPA beyond class-conditional synthesis, we conduct text-to-image (T2I) generation experiments on the MS-COCO dataset \cite{lin2014MSCOCO}. This task requires the model to capture complex cross-modal alignments between free-form text and visual structures. Following the established protocol \cite{yu2024repa, bao2023U-ViT}, we adopt MMDiT \cite{esser2024MMDiT} as the diffusion backbone, which jointly processes image patches and text embeddings.

\textbf{Experimental Setup.} We train the MMDiT model from scratch on the MS-COCO training split for 150K iterations with a total batch size of 256. The model configuration consists of a hidden dimension of 768 and a depth of 24 layers. Text prompts are derived using a pre-trained CLIP text encoder \cite{radford2021CLIP}. During inference, we evaluate performance on the validation split with a classifier-free guidance scale of 2.0, reporting FID under both ODE (NFE=50) and SDE (NFE=250) samplers.

\textbf{Performance Analysis.} Quantitative results are summarized in Table~\ref{tab:t2i_results}. AHPA achieves a superior FID of 4.52 (SDE) and 5.02 (ODE), outperforming the previous self-alignment SOTA, SRA 2, by a significant margin. Remarkably, despite relying solely on intrinsic VAE features, AHPA matches or even slightly surpasses REPA, which utilizes heavy-weight external vision learners (DINOv2). These results indicate that the hierarchical priors recovered by AHPA, when adaptively scheduled, provide sufficient semantic-spatial guidance for complex text-conditioned tasks. The consistency of these gains across different samplers further underscores the robustness of our dynamic routing mechanism in resolving representational mismatch.

\begin{table}[t]
\centering
\caption{\textbf{Quantitative comparison on text-to-image generation (MS-COCO)}. Following the established experimental protocol, we use classifier-free guidance with $w = 2.0$ for all evaluations. AHPA achieves competitive performance against methods relying on external representation learners, despite using only intrinsic VAE features.}
\label{tab:t2i_results}
\small
\begin{tabularx}{\columnwidth}{@{\extracolsep{\fill}} l c c @{}}
\toprule
Method & Type & FID $\downarrow$ \\
\midrule
AttnGAN \cite{xu2017AttnGAN} & GAN & 35.49 \\
DM-GAN \cite{zhu2019DM-GAN} & GAN & 32.64 \\
VQ-Diffusion \cite{gu2022VQ-Diffusion} & Discrete Diffusion & 19.75 \\
DF-GAN \cite{tao2022DF-GAN} & GAN & 19.32 \\
XMC-GAN \cite{zhang2022XMC-GAN} & GAN & 9.33 \\
Frido \cite{fan2022frido} & Diffusion & 8.97 \\
LAFITE \cite{zhou2022lafite} & GAN & 8.12 \\
\midrule
U-Net \cite{bao2023U-Net} & Diffusion & 7.32 \\
U-ViT-S/2 \cite{bao2023U-ViT} & Diffusion & 5.95 \\
U-ViT-S/2 (Deep) \cite{bao2023U-ViT} & Diffusion & 5.48 \\
\midrule
MMDiT \cite{esser2024MMDiT} (ODE; NFE=50) & Diffusion & 6.05 \\
MMDiT+REPA \cite{yu2024repa} (ODE; NFE=50) & Diffusion & 5.37 \\
MMDiT+SRA 2 \cite{wang2026sra2variationalautoencoder} (ODE; NFE=50)& Diffusion & 5.65 \\
\rowcolor{gray!20} \textbf{MMDiT+AHPA (ODE; NFE=50)} & Diffusion & \textbf{5.02} \\
\midrule
MMDiT \cite{esser2024MMDiT} (SDE; NFE=250) & Diffusion & 5.30 \\
MMDiT+REPA \cite{yu2024repa} (SDE; NFE=250) & Diffusion & 4.74 \\
MMDiT+SRA 2 \cite{wang2026sra2variationalautoencoder} (SDE; NFE=250)& Diffusion & 4.91 \\
\rowcolor{gray!20} \textbf{MMDiT+AHPA (SDE; NFE=250)} & Diffusion & \textbf{4.52} \\
\bottomrule
\end{tabularx}
\end{table}
\section{Additional Ablation Studies and Technical Details} \label{app:additional_ablations}
\subsection{Comprehensive Component Ablation}
\label{app:component_ablation}

To identify the optimal configuration for AHPA, we conduct an extensive ablation study on the projection head architecture, the alignment loss function, and the supervision weight $\lambda$. 

\textbf{Analysis of Results.} As summarized in Table~\ref{tab:comprehensive_ablation}, our findings indicate several key design principles:
\begin{itemize}
    \item \textbf{Loss Function:} \textbf{Cosine Similarity} outperforms L1-based losses, suggesting that maintaining directional consistency in the diffusion manifold is more vital than regressing absolute magnitudes.
    \item \textbf{Weighting Strategy:} $\lambda=1.0$ provides the optimal balance; higher weights (e.g., $\lambda=2.0$) slightly enhance structural metrics but risk distorting the generative manifold and compromising image realism.
    \item \textbf{Convolutional Advantage:} Single-layer $3\times3$ Conv significantly outperforms MLP and Linear heads. By leveraging spatial inductive bias, the convolutional head effectively captures local structural dependencies, smoothing resampling artifacts for superior alignment quality.
    \item \textbf{Minimalism vs. Depth:} Stacking deeper blocks (e.g., Conv + GN + SiLU) leads to performance degradation. This indicates that over-processing hierarchical priors via redundant non-linearities distorts their intrinsic distribution, whereas a single-layer conv preserves the highest guidance fidelity.
\end{itemize}

\begin{table}[h]
\centering
\caption{\textbf{Component-wise ablation study of AHPA.} All experiments are conducted on SiT-B/2 for 400K iterations. We evaluate the impact of projection head architecture, loss functions, and alignment weight $\lambda$. The configuration (l) represents the default setting used in AHPA.}
\label{tab:comprehensive_ablation}
\small
\setlength{\tabcolsep}{12pt}
\begin{tabular}{l c c c c c}
\toprule
\textbf{Variant} & \textbf{Proj. Head} & \textbf{Loss} & $\lambda$ & \textbf{FID} $\downarrow$ & \textbf{sFID} $\downarrow$ \\
\midrule
(a) Vanilla SiT & --- & --- & --- & 33.0 & 6.46 \\
\midrule
\textit{Loss \& Weight} & & & & & \\
(b) & 1-layer $3\times3$ Conv & L1 & 1.0 & 27.2 & 6.12 \\
(c) & 1-layer $3\times3$ Conv & Smooth-L1 & 1.0 & 26.9 & 5.95 \\
(d) & 1-layer $3\times3$ Conv & Cosine & 0.5 & 27.8 & 6.30 \\
(e) & 1-layer $3\times3$ Conv & Cosine & 2.0 & 26.4 & 5.82 \\
\midrule
\textit{Architecture Depth} & & & & & \\
(f) & 1-layer Linear & Cosine & 1.0 & 28.2 & 6.45 \\
(g) & 3-layer MLP & Cosine & 1.0 & 26.8 & 6.10 \\
(h) & 5-layer MLP & Cosine & 1.0 & 27.5 & 6.35 \\
(i) & 2-layer Conv Block & Cosine & 1.0 & 26.2 & 6.01 \\
(j) & 3-layer Conv Block & Cosine & 1.0 & 27.4 & 6.22 \\
(k) & 5-layer Conv Block & Cosine & 1.0 & 28.1 & 6.41 \\
\midrule
\rowcolor{highlightpink} \textbf{(l)} & \textbf{1-layer $3\times3$ Conv} & \textbf{Cosine} & \textbf{1.0} & \textbf{25.5} & \textbf{5.78} \\
\bottomrule
\end{tabular}
\end{table}
\subsection{Sensitivity to Hierarchical Partitioning Boundary}
\label{app:partition_sensitivity}

To address the potential ambiguity in partitioning VAE layers with identical channel dimensions ($C=512$), we evaluate the model's sensitivity to the boundary between $\mathcal{G}_{mid}$ and $\mathcal{G}_{deep}$. Specifically, while keeping the total alignment depth constant at 8 layers, we vary the number of layers assigned to each group. 

As shown in Table~\ref{tab:partition_sensitivity}, the balanced configuration ($4+4$) achieves the optimal FID, though the performance remains relatively stable across different ratios. This stability suggests that AHPA's effectiveness stems from the functional gradient inherent in the VAE hierarchy rather than a hyper-tuned split point. The slight degradation in the $6+2$ setting indicates that insufficient semantic supervision may weaken global composition, while the $2+6$ setting slightly hampers structural precision. Consequently, we adopt the balanced split as a robust and parameter-free default.

\begin{table}[h]
    \centering
    \caption{\textbf{Sensitivity analysis of the partitioning boundary.} We vary the layer distribution between  $\mathcal{G}_{mid}$ and $\mathcal{G}_{deep}$ while maintaining a total budget of 8 layers. Evaluated on SiT-B/2 at 400K iterations.}
    \label{tab:partition_sensitivity}
    \small
    \setlength{\tabcolsep}{10pt}
    \begin{tabular}{cc|ccc}
        \toprule
        $\mathcal{G}_{mid}$ Layers & $\mathcal{G}_{deep}$ Layers & FID $\downarrow$ & sFID $\downarrow$ & IS $\uparrow$ \\
        \midrule
        2 & 6 & 26.3 & 5.95 & 50.8 \\
        \rowcolor{gray!10} \textbf{4 (Default)} & \textbf{4 (Default)} & \textbf{25.5} & \textbf{5.78} & \textbf{52.3} \\
        6 & 2 & 27.1 & 6.12 & 50.5 \\
        \bottomrule
    \end{tabular}
\end{table}
\subsection{Extended Analysis of the Non-stationary Alignment Requirement}

The efficacy of AHPA stems from its ability to synchronize guidance granularity with the model's evolving representational needs. To isolate the contribution of the routing mechanism, we contrast our Dynamic Router against two baseline strategies in Table \ref{tab:dynamic_vs_static}:

\textbf{Representational vs. Scheduling Gains.} The transition from Vanilla SiT-B to Static Uniform alignment provides a 3.3 FID improvement (33.0 $\to$ 29.7), establishing hierarchical VAE features as a robust foundation. However, a fixed linear Heuristic schedule yields only a marginal 1.1 FID gain (29.7 $\to$ 28.6). This limited improvement suggests that a simplistic linear transition is insufficient to capture the complex, non-stationary dynamics of the diffusion trajectory.

\textbf{Autonomy vs. Manual Heuristics.} In contrast, our learned Dynamic Router achieves a substantial 4.2 FID gain over the static baseline (29.7 $\to$ 25.5), outperforming the linear heuristic by a significant 3.1 FID margin. This gap confirms that the granularity-noise tradeoff requires precise, non-linear coordination that emerges naturally through our adaptive mechanism. 

Crucially, while one might attempt to design a hand-tuned non-linear schedule (e.g., a Sigmoid function) based on the diagnostic crossovers in Sec \ref{sec:mechanics}, such closed-form solutions are inherently fragile. Our observations indicate that the optimal crossover point is architecture-dependent; for instance, the $t \approx 0.4$ crossover identified for SiT-XL does not necessarily generalize to SiT-B. The primary advantage of our router is its autonomy---it eliminates the need for expensive per-model diagnostic probing and manual parameter tuning, providing a generalized solution that automatically discovers the optimal scheduling curvature for any given architecture.

\begin{table}[h]
    \centering
    \caption{\textbf{Necessity of Learned Dynamic Routing.} Evaluated on SiT-B/2. We contrast our learned router against fixed weighting and a manually designed linear scheduling baseline (Heuristic). Our adaptive mechanism consistently maximizes guidance efficiency.}
    \label{tab:dynamic_vs_static}
    {\small
    \begin{tabular}{llccc}
        \toprule
        Feature Source & Strategy & IS $\uparrow$ & FID $\downarrow$ & Prec. $\uparrow$ \\
        \midrule
        Vanilla SiT-B & -- & 43.7 & 33.0 & 0.53 \\
        \midrule
        \multirow{3}{*}{$G_3 \cup G_4$} 
        & Static Uniform & 51.6 & 29.7 & 0.58 \\
        & Heuristic (Linear) & 50.2 & 28.6 & 0.59 \\
        \cmidrule{2-5}
        \rowcolor{gray!10}
        & \textbf{Dynamic Routing (Ours)} & \textbf{52.3} & \textbf{25.5} & \textbf{0.63} \\
        \bottomrule
    \end{tabular}
    }
\end{table}
\subsection{Positional Optimality and Proportional Scaling}
A critical design decision in AHPA is the choice of the alignment block $l$. In our preliminary study using static guidance, we identified block $l=3$ as the optimal entry point for SiT-B. To ensure this optimality persists under the dynamic regime, we evaluate AHPA at various alternative blocks.

As shown in Table \ref{tab:ahpa_layer_verification}, while the Dynamic Router improves performance across all tested blocks compared to static baselines, the peak consistently remains at $l=3$. This reinforces the observation that the early-to-mid stages of the Transformer backbone are uniquely receptive to hierarchical priors. At these depths, the model has begun to aggregate global context but has not yet committed to the fine-grained pixel details that characterize the final blocks. 

Consequently, we adopt a proportional scaling strategy across model sizes. By placing the alignment at approximately the 25th percentile of the total depth (block 3 for Base, 6 for Large, and 7 for XL), we ensure that AHPA consistently targets the most "flexible" representational phase of the backbone, facilitating efficient knowledge transfer from the VAE's hierarchical blueprint.

\begin{table}[h]
    \centering
    \caption{\textbf{Verification of positional optimality under AHPA}. We evaluate our dynamic mechanism at different blocks to confirm if the "golden entry point" identified in the static setting ($l=3$) remains optimal even with adaptive scheduling.}
    \label{tab:ahpa_layer_verification}
    \begin{small}
    \begin{tabular}{lccc}
        \toprule
        Alignment Block $l$ & \textbf{3} & 4 & 8 \\
        \midrule
        Strategy & \textbf{Dynamic} & Dynamic & Dynamic \\
        FID $\downarrow$ & \textbf{25.5} & 27.2 & 31.5 \\
        IS $\uparrow$ & \textbf{52.3} & 50.8 & 46.4 \\
        \bottomrule
    \end{tabular}
    \end{small}
\end{table}

\subsection{Hierarchical Conflict and Optimization Stability}
Further analysis of the "Full Union (G1--G4)" experiment in Table \ref{tab:group_ablation} provides insight into why localized guidance (G3 $\cup$ G4) is superior. Shallow VAE features (G1, G2) are tightly coupled with the input pixel space. When forced as alignment targets, they compel the Transformer's early layers to reconstruct low-level details prematurely. This creates an optimization bottleneck, as the model's capacity is diverted from learning the generative manifold toward simple pixel replication. By excluding these shallow layers, AHPA focuses the alignment signal on the mid-to-deep features that capture the semantic and structural feature of the image, leading to faster convergence and higher fidelity.
\subsection{Diversified exploration of Dynamic Router Design}
\label{app:router_detailed_ablation}

To provide a thorough understanding of the Dynamic Router $\mathcal{R}_\phi$, we ablate its design across four critical dimensions. These experiments justify why a minimalist, timestep-conditioned MLP is the most effective architecture for coordinating hierarchical priors. All models are SiT-B/2 scales trained for 400K iterations.

\textbf{1. Timestep Encoding Strategy.}
We evaluate how the representation of scalar $t \in [0, 1]$ affects the quality of manifold alignment. Formally, we compare three encoding schemes:
\begin{itemize}
    \item \textbf{Linear Projection (Ours):} Defined as $e_{linear}(t) = \mathbf{W}t + \mathbf{b}$, where $\mathbf{W} \in \mathbb{R}^{D \times 1}$ and $\mathbf{b} \in \mathbb{R}^D$ are learnable. The constant Jacobian $\frac{\partial e}{\partial t} = \mathbf{W}$ ensures that the router's input changes at a uniform rate, preserving the monotonic nature of the diffusion trajectory and facilitating a seamless transition between different stages.
    \item \textbf{Sinusoidal Embedding:} Following \cite{ho2020ddpm}, $e_{sin}(t) = [ \dots, \sin(\omega_i t), \cos(\omega_i t), \dots ]^\top$ with $\omega_i = 10000^{-2i/D}$. While standard for feature conditioning, its high-frequency harmonic components introduce periodic local fluctuations in the weights $\alpha(t)$, which destabilizes the alignment objective (27.0 FID).
    \item \textbf{Fourier Features (RFF):} Maps $t$ via $e_{RFF}(t) = [ \dots, \cos(2\pi \mathbf{b}_i t), \sin(2\pi \mathbf{b}_i t), \dots ]^\top$ where $\mathbf{b}_i \sim \mathcal{N}(0, \sigma^2)$. This leads to "weight aliasing," where the router over-fits to micro-variations in noise levels, yielding a degraded FID of 26.8.
\end{itemize}

\textbf{2. Informational Source: Global Sync vs. Local Adaptation.}
We evaluate whether the routing policy $\phi$ should depend on the global denoising schedule or adapt to localized feature representations. We contrast our timestep-only design with a content-adaptive variant.

\textit{Implementation of Content Feature:} To capture the specific state of the model at each alignment step, the Content-adaptive router extracts the intermediate feature map $h^{(l)} \in \mathbb{R}^{C \times H \times W}$ directly from the target Transformer layer being aligned. We apply global average pooling to collapse the spatial dimensions, resulting in a compact representational vector $\bar{h} \in \mathbb{R}^C$. This vector is then projected via a linear layer and fed into the router MLP. For a fair comparison, the Content-adaptive variant also receives the timestep embedding $e(t)$ to ensure it has access to the full temporal context.

\textit{Failure Analysis:} As shown in Table~\ref{tab:router_ablation}, the Content-adaptive approach significantly degrades performance (29.6 vs. 25.5 FID). We attribute this to the stochastic noise dominance in early denoising stages (high $t$). Attempting to derive routing policies from these noise-heavy features introduces high-frequency "jitter" into the hierarchical weights $\alpha(t)$, preventing the backbone from establishing a consistent representational gradient. This confirms that the demand for VAE priors is a trajectory-wide property governed by global SNR evolution, rather than a local instance-specific one.

\textbf{3. Router Architecture and Complexity: From Discreteness to Continuity.} 
We contrast our MLP-based $\mathcal{R}_\phi$ with two alternative structural paradigms to verify the necessity of continuous non-linear mapping for inter-group ($\beta$) and intra-group ($\alpha$) scheduling.

\begin{itemize}
    \item \textbf{Lookup Table (LUT):} 
    \textit{Implementation:} We discretize the continuous timestep $t \in [0, 1]$ into $N=10$ uniform bins. The router is parameterized by two learnable matrices, $\mathbf{M}_{\beta} \in \mathbb{R}^{N \times G}$ and $\mathbf{M}_{\alpha} \in \mathbb{R}^{N \times (G \times L)}$. For any input $t$, the index $idx = \lfloor t \cdot N \rfloor$ is used to retrieve the corresponding weighting vectors:
    $$\begin{bmatrix} \beta(t) \\ \alpha(t) \end{bmatrix} = \text{Lookup}(\mathbf{M}, idx)$$
    \textit{Failure Analysis:} This formulation creates a piecewise constant function. As $t$ transitions across a bin boundary (e.g., from 0.899 to 0.901), the routing weights undergo an instantaneous step change. Such discontinuities act as high-frequency "gradient shocks" to the backbone, forcing the transformer to abruptly adapt to new alignment scales. This disrupts the smooth evolution of the latent manifold, resulting in structural artifacts and a degraded FID of 27.2.

    \item \textbf{Attention-based Router:}
    \textit{Implementation:} We implement a dynamic retrieval mechanism using a Query-Key-Value (QKV) framework. We define a set of $K=16$ learnable \textit{hierarchical prototypes} $\{\mathbf{k}_i, \mathbf{v}_i\}_{i=1}^K$, where each $\mathbf{v}_i$ represents a potential $(\alpha, \beta)$ configuration. Given the timestep embedding $e(t)$, the weights are generated via:
    $$w_i = \frac{\exp(e(t) \cdot \mathbf{k}_i / \sqrt{D})}{\sum_{j=1}^K \exp(e(t) \cdot \mathbf{k}_j / \sqrt{D})}, \quad \begin{bmatrix} \beta(t) \\ \alpha(t) \end{bmatrix} = \sum_{i=1}^K w_i \cdot \mathbf{v}_i$$
    \textit{Failure Analysis:} While highly expressive, the Attention-based router is over-engineered for the alignment task. The redundant complexity of the attention bank introduces unnecessary optimization noise and potential over-fitting to specific timesteps, yielding a sub-optimal FID of \textbf{26.9}.
\end{itemize}

\textit{Conclusion:} Our 4-layer MLP provides the ideal inductive bias for this task. Unlike LUT, it ensures $\mathcal{C}^\infty$ continuity across the entire trajectory; unlike Attention, it focuses on global functional mapping rather than discrete pattern retrieval. This leads to the most stable alignment signal and the best generative performance (25.5 FID).

\textbf{4. Depth of the MLP.}
We sweep the number of layers $L \in \{3, 4, 8\}$ for our MLP router. A 3-layer MLP lacks the non-linearity required for the sharp semantic-to-structural transition (26.1 FID). We find that $L=4$ provides the optimal balance; further increasing the depth to $L=8$ results in inferior performance (26.3 FID), suggesting that excessive complexity may hinder the optimization of the temporal routing policy.

\begin{table}[h]
\centering
\caption{\textbf{Ablation Study on Router Configurations.} We systematically evaluate the impact of encoding strategies, input sources, and architectural depths on SiT-B/2 for 400K iterations. Each row represents a specific experimental configuration. \textbf{Bold} indicates our final optimized setting.}
\label{tab:router_ablation}
\small
\setlength{\tabcolsep}{8pt}
\begin{tabular}{c l l l c c}
\toprule
Exp. & \textit{t} Encoding & Input Source & Architecture & MLP Depth & FID $\downarrow$ \\
\midrule
1 & Sinusoidal & Timestep-only & MLP & 4 & 27.0\\
2 & Fourier (RFF) & Timestep-only & MLP & 4 & 26.8 \\
3 & Linear & Content-adaptive & MLP & 4 & 29.6 \\
4 & Linear & Timestep-only & Lookup Table & --- & 27.2 \\
5 & Linear & Timestep-only & Attention-based & --- & 26.9 \\
\midrule
6 & Linear & Timestep-only & MLP & 3 & 26.1\\
\rowcolor{highlightpink} \textbf{7} & \textbf{Linear} & \textbf{Timestep-only} & \textbf{MLP} & \textbf{4} & \textbf{25.5} \\
8 & Linear & Timestep-only & MLP & 8 & 26.3\\
\bottomrule
\end{tabular}
\end{table}

\section{Theoretical Grounding and Validity of G-SNR} \label{app:metric_validation}

To provide a rigorous justification for the Gradient Signal-to-Noise Ratio (G-SNR), we situate this metric within the dual framework of multi-task optimization and information propagation. We demonstrate that G-SNR is a principled measure of \textit{gradient stationarity} and \textit{representational receptivity}.

\subsection{G-SNR: Quantifying Gradient Purity and Task Conflict}

The AHPA objective, $\mathcal{L}_{total} = \mathcal{L}_{diff} + \lambda \mathcal{L}_{align}$, constitutes a multi-task optimization problem. According to the theory of \textit{Gradient Surgery} \cite{yu2020Gradient}, the efficacy of joint optimization is governed by the directional alignment of task-specific gradients. If $\cos(g_{diff}, g_{align}) < 0$, the objectives exhibit gradient conflict, where alignment updates counteract the manifold convergence of the denoising task.

\paragraph{Statistical Estimation via Per-sample Gradients.} 
To provide a robust diagnostic independent of optimization hyperparameters, we estimate G-SNR using the statistical properties of the gradient vector field within each mini-batch. Let $\mathcal{B} = \{x_1, \dots, x_B\}$ be a mini-batch of size $B$, and $g_t^{(i)} = \nabla_\theta \mathcal{L}_{align}(x_i, t)$ be the alignment gradient elicited by the $i$-th individual sample. We define the empirical G-SNR estimator as:

\begin{equation}
    \widehat{\text{G-SNR}}(t) = \frac{\left\| \frac{1}{B} \sum_{i=1}^B g_t^{(i)} \right\|_2^2}{\frac{1}{B-1} \sum_{i=1}^B \left\| g_t^{(i)} - \bar{g}_t \right\|_2^2}
\end{equation}

where $\bar{g}_t = \frac{1}{B} \sum_{i=1}^B g_t^{(i)}$ is the mini-batch average gradient. 

\paragraph{Physical Interpretation and Robustness.} 
Physically, G-SNR measures signal purity: a high G-SNR signifies that the hierarchical prior provides a \textit{stationary} signal that resides within the constructive optimization subspace of the denoising manifold. Conversely, a low G-SNR indicates that $g_{align}$ is dominated by stochastic variance, effectively ``poisoning'' the trajectory with impurity noise. 

Notably, this formulation renders the metric dimensionless and invariant to absolute gradient magnitudes. Since both the numerator (squared expectation) and the denominator (trace of the covariance matrix) scale quadratically with the gradient magnitude, G-SNR reflects directional consistency rather than the absolute energy of the update. This scale-invariance is critical for assessing alignment health across the non-stationary diffusion trajectory where gradient scales vary significantly.
\subsection{G-SNR as a Proxy for Representational Receptivity}
Beyond stability, G-SNR implicitly measures the backbone's \textit{receptivity} to specific priors. As established in \textit{Signal Propagation} theory \cite{saxe2014Exact}, a model's ability to extract information from a supervisor is limited by its instantaneous receptive field and abstraction level. A diminishing G-SNR (e.g., for Mid-level priors at $t > 0.8$) signifies a state of representational ``blindness,'' where the backbone lacks the geometric maturity to resolve the prior, causing the gradient to dissipate into unconstructive variance. By maximizing G-SNR, AHPA ensures that guidance is injected only when the model has the capacity to internalize it.

\subsection{Alignment with Diffusion SNR Physics}
Crucially, G-SNR is coupled with the non-stationary physics of diffusion. \textit{Variational Diffusion Models} \cite{kingma2023Variational} show that the focus transitions from global anchoring to local refinement as input SNR increases. G-SNR acts as the representation-space counterpart to this schedule. The crossover behaviors identified in our diagnostic provide mechanistic evidence that the demand for prior ``purity'' is synchronized with the noise schedule, justifying our use of a Dynamic Router.

\subsection{Empirical Validity and Interpretability}
We validate G-SNR as a diagnostic tool by monitoring its evolution across methods. As shown in Table~\ref{tab:app_consistency}, there is a high degree of consistency between G-SNR and generative performance. Methods that suffer from low trajectory-average G-SNR (e.g., Uniform $\mathcal{G}_{deep}$ and SRA 2) consistently exhibit inferior structural fidelity. Notably, AHPA achieves the highest Avg. G-SNR, which mechanistically aligns with its superior sFID and FID scores. This suggests that G-SNR is a \textit{reasonable and robust diagnostic metric} for assessing alignment health, offering insights into signal purity that standard loss curves fail to capture.

\begin{table}[h]
\centering
\caption{\textbf{Comparison of diagnostic health and generative performance.} Evaluated on SiT-XL/2 at 400K iterations (ImageNet-256). Trajectory Avg. G-SNR serves as a mechanism-consistent indicator of the alignment process's health, aligning with the improvements observed in structural fidelity (sFID).}
\label{tab:app_consistency}
\small
\setlength{\tabcolsep}{12pt}
\begin{tabular}{lccc}
\toprule
Method & FID $\downarrow$ & sFID $\downarrow$ & \textbf{Avg. G-SNR} $\uparrow$ \\
\midrule
Uniform $\mathcal{G}_{deep}$ & 11.2 & 5.36 & 7.3 \\
Uniform $\mathcal{G}_{mid}$  & 14.1 & 6.25 & 7.0 \\
SRA 2 (Latent)             & 13.8 & 5.10 & 7.2 \\
REPA (DINOv2-B)            & 8.5  & 5.14 & 8.5 \\
\midrule
\rowcolor{highlightpink} \textbf{AHPA (Ours)} & \textbf{8.1} & \textbf{4.90} & \textbf{11.6} \\
\bottomrule
\end{tabular}
\end{table}
\section{512$\times$512 ImageNet Generation}
\label{512comparison}
To evaluate the scalability and high-resolution synthesis capabilities of our method, we conduct experiments on the ImageNet $512\times512$ dataset using the SiT-XL/2 backbone. 

As shown in Table~\ref{tab:512performance}, AHPA maintains its efficiency lead on $512\times512$ resolution, where it achieves an FID of 2.28 in only 100 epochs—already outperforming the vanilla SiT-XL/2 baseline trained for the full 600 epochs. Its substantial lead in sFID (4.16) confirms that AHPA preserves semantic structures more effectively than the vanilla SiT and static alignment strategies.
\begin{table}[h]
\centering
\caption{Performance comparison on ImageNet $512\times512$ with CFG.}
\label{tab:512performance}
\small
\begin{tabular}{lcccccc}
\toprule
Model & Epochs & FID$\downarrow$ & sFID$\downarrow$ & IS$\uparrow$ & Pre.$\uparrow$ & Rec.$\uparrow$ \\ 
\midrule
\multicolumn{7}{l}{\textit{Pixel diffusion}} \\
VDM++ \cite{kingma2023VDM++} & - & 2.65 & - & 278.1 & - & - \\
ADM-G, ADM-U \cite{dhariwal2021ADM-U} & 400 & 2.85 & 5.86 & 221.7 & 0.84 & 0.53 \\
Simple diffusion (U-Net) \cite{bao2023U-Net} & 800 & 4.28 & - & 171.0 & - & - \\
Simple diffusion (U-ViT, L) \cite{bao2023U-ViT} & 800 & 4.53 & - & 205.3 & - & - \\
\midrule
\multicolumn{7}{l}{\textit{Latent diffusion, Transformer}} \\
MaskDiT \cite{zheng2024maskdit} & 800 & 2.50 & 5.10 & 256.3 & 0.83 & 0.56 \\
\midrule
DiT-XL/2 \cite{peebles2023scalable} & 600 & 3.04 & 5.02 & 240.8 & 0.84 & 0.54 \\
\midrule
SiT-XL/2 \cite{ma2024sit} & 600 & 2.62 & 4.18 & 252.2 & \textbf{0.84} & \textbf{0.57} \\
+ SRA 2 \cite{wang2026sra2variationalautoencoder}& 100 & 2.45 & 4.28 & 249.1 & 0.83 & 0.55 \\
+ REPA \cite{yu2024repa}& 100 & 2.37 & 4.20 & 253.7 & 0.83 & 0.56 \\
\rowcolor{highlightpink} 
\textbf{+ AHPA (ours)} & 100 & \textbf{2.28} & \textbf{4.16} & \textbf{261.4} & \textbf{0.84} & 0.56 \\
\bottomrule
\end{tabular}
\end{table}
\section{Detailed Comparison with SOTA Methods}
Detailed comparisons with SOTA methods are provided in Table \ref{tab:whole_sota_compare}.
\begin{table*}[t]
\centering
\caption{\textbf{State-of-the-Art Comparison on ImageNet $256\times256$ with CFG.} Performance metrics are annotated with $\uparrow$ (higher is better) and $\downarrow$ (lower is better). \textbf{w/o Ext.} indicates whether the method is standalone and does \textit{not} rely on external dependencies (e.g., extra encoders, diffusion teachers, or diffusion decoders). Bold indicates the best results among methods without external dependencies.}
\label{tab:whole_sota_compare}
\footnotesize 
\setlength{\tabcolsep}{2pt} 
\begin{tabularx}{\textwidth}{@{\extracolsep{\fill}} lccccccc @{}}
\toprule
Method & Epochs & FID$\downarrow$ & sFID$\downarrow$ & IS$\uparrow$ & Pre.$\uparrow$ & Rec.$\uparrow$ & w/o Ext. \\ \midrule
\textit{Pixel diffusion} & & & & & & & \\
ADM-U \cite{dhariwal2021ADM-U} & 400 & 3.94 & 6.14 & 186.7 & 0.82 & 0.52 & \greenCheck \\
VDM++ \cite{kingma2023VDM++} & 560 & 2.40 & - & 225.3 & - & - & \greenCheck \\
Simple diffusion \cite{hoogeboom2023simplediffusion} & 800 & 2.77 & - & 211.8 & - & - & \greenCheck \\
CDM \cite{ho2021CDM} & 2160 & 4.88 & - & 158.7 & - & - & \greenCheck \\ \midrule
\textit{Latent diffusion, U-Net} & & & & & & & \\
LDM-4 \cite{rombach2022LDM-4}& 200 & 3.60 & - & 247.7 & 0.87 & 0.48 & \greenCheck \\ \midrule
\textit{Latent diffusion, Transformer + U-Net hybrid} & & & & & & & \\
U-ViT-H/2 \cite{bao2023U-ViT} & 240 & 2.29 & 5.68 & 263.9 & 0.82 & 0.57 & \greenCheck \\
DiffiT \cite{hatamizadeh2024Diffit} & - & 1.73 & - & 276.5 & 0.80 & 0.62 & \greenCheck \\ \midrule
\textit{Latent diffusion, Transformer} & & & & & & & \\
DiT-XL/2 \cite{peebles2023scalable} & 1400 & 2.27 & 4.60 & 278.2 & \textbf{0.83} & 0.57 & \greenCheck \\
+ iSSD \cite{Ma2025iSSD} & 1400+0.8 & 2.02 & 4.22 & 250.0 & 0.81 & 0.60 & \redCross \\
MaskDiT \cite{zheng2024maskdit} & 1600 & 2.28 & 5.67 & 276.6 & 0.80 & 0.61 & \redCross \\
SD-DiT \cite{zhu2024SD-DiT} & 480 & 3.23 & - & - & - & - & \redCross \\ \addlinespace
SiT-XL/2 \cite{ma2024sit} & 1400 & 2.06 & 4.50 & 270.3 & 0.82 & 0.59 & \greenCheck \\
+ REPA \cite{yu2024repa} & 800 & 1.42 & 4.70 & 311.4 & 0.80 & 0.63 & \redCross \\
+ REG \cite{wu2025reg} & 800 & 1.36 & 4.25 & 299.4 & 0.77 & 0.66 & \redCross \\
+ SRA \cite{jiang2026sra} & 800 & 1.58 & 4.65 & 305.7 & 0.80 & 0.63 & \greenCheck \\
+ SRA 2 \cite{wang2026sra2variationalautoencoder} & 800 & 1.52 & 4.63 & 316.2 & 0.82 & 0.62 & \greenCheck \\ \addlinespace
\rowcolor{highlightpink} \cellcolor{white} + \textbf{AHPA (Ours)} & 100 & 2.15 & 4.38 & 265.4 & 0.81 & 0.58 & \greenCheck \\
\rowcolor{highlightpink} \cellcolor{white} + \textbf{AHPA (Ours)} & 200 & 1.78 & 4.26 & 291.8 & 0.82 & 0.61 & \greenCheck \\
\rowcolor{highlightpink} \cellcolor{white} + \textbf{AHPA (Ours)} & 400 & 1.56 & 4.15 & 308.2 & 0.81 & 0.63 & \greenCheck \\
\rowcolor{highlightpink} \cellcolor{white} + \textbf{AHPA (Ours)} & 800 & \textbf{1.40} & \textbf{4.12} & \textbf{317.5} & 0.81 & \textbf{0.64} & \greenCheck \\ \bottomrule
\end{tabularx}
\end{table*}
\section{Qualitative Comparison across Different Methods}
\label{app:qualitative_comparison}
\begin{figure*}[!t]
  \centering
  \small
  \resizebox{0.95\textwidth}{!}{
    \setlength{\tabcolsep}{0pt}  
    \begin{tabular}{@{} l @{\hspace{8pt}} ccc @{\hspace{12pt}} ccc @{}}
      & 100K & 200K & 400K & 100K & 200K & 400K \\
      \addlinespace[2pt]
      
      \adjustbox{valign=b}{\rotatebox{90}{\scriptsize \qquad \quad SRA 2}} &
      \includegraphics[width=0.15\textwidth]{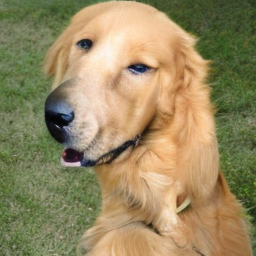} &
      \includegraphics[width=0.15\textwidth]{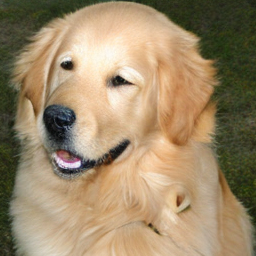} &
      \includegraphics[width=0.15\textwidth]{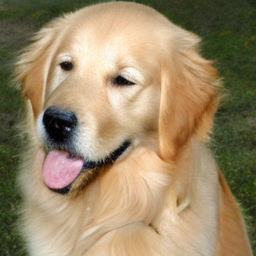} &
      \includegraphics[width=0.15\textwidth]{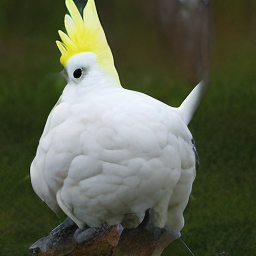} &
      \includegraphics[width=0.15\textwidth]{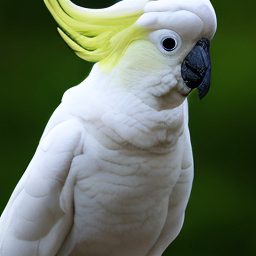} &
      \includegraphics[width=0.15\textwidth]{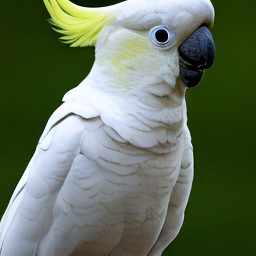} \\
      
      \adjustbox{valign=b}{\rotatebox{90}{\scriptsize \qquad \quad REPA}} &
      \includegraphics[width=0.15\textwidth]{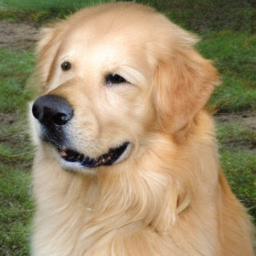} &
      \includegraphics[width=0.15\textwidth]{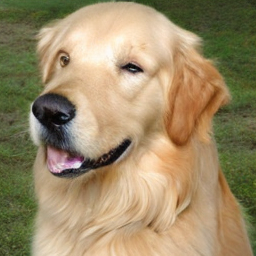} &
      \includegraphics[width=0.15\textwidth]{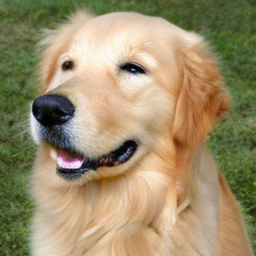} &
      \includegraphics[width=0.15\textwidth]{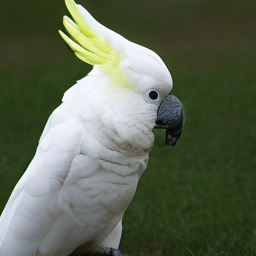} &
      \includegraphics[width=0.15\textwidth]{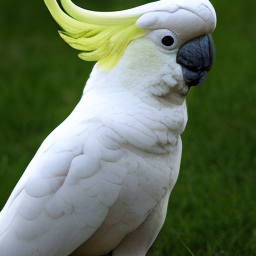} &
      \includegraphics[width=0.15\textwidth]{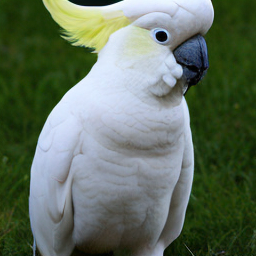} \\
      
      \adjustbox{valign=b}{\rotatebox{90}{\scriptsize \qquad \quad AHPA}} &
      \includegraphics[width=0.15\textwidth]{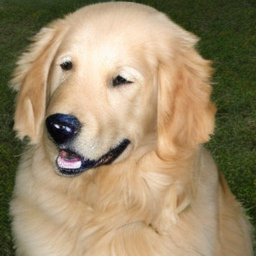} &
      \includegraphics[width=0.15\textwidth]{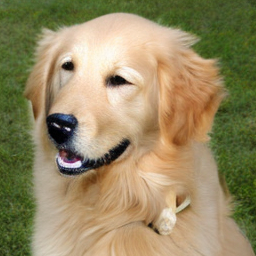} &
      \includegraphics[width=0.15\textwidth]{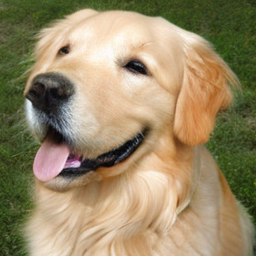} &
      \includegraphics[width=0.15\textwidth]{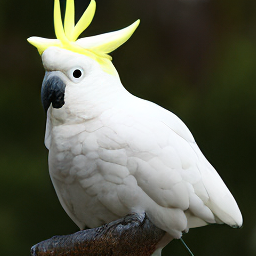} &
      \includegraphics[width=0.15\textwidth]{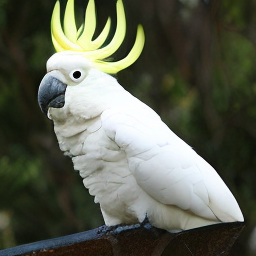} &
      \includegraphics[width=0.15\textwidth]{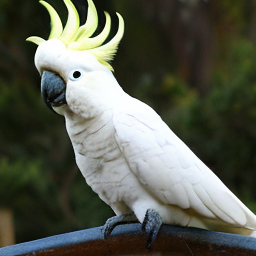} \\
      
      \addlinespace[4pt] 

      
      \adjustbox{valign=b}{\rotatebox{90}{\scriptsize \qquad \quad SRA 2}} &
      \includegraphics[width=0.15\textwidth]{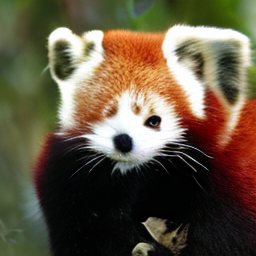} &
      \includegraphics[width=0.15\textwidth]{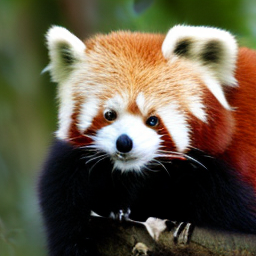} &
      \includegraphics[width=0.15\textwidth]{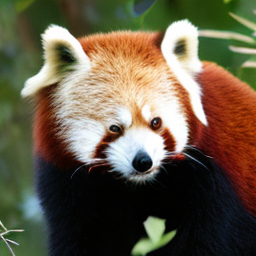} &
      \includegraphics[width=0.15\textwidth]{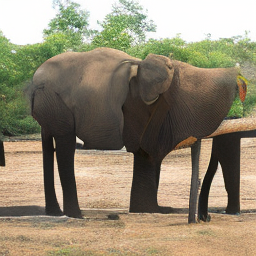} &
      \includegraphics[width=0.15\textwidth]{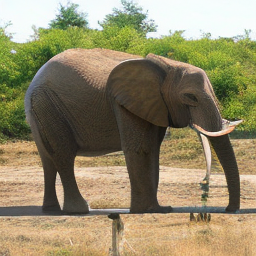} &
      \includegraphics[width=0.15\textwidth]{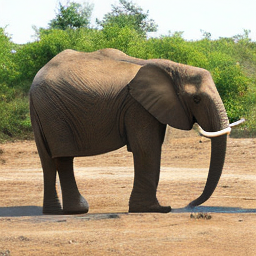} \\
      
      \adjustbox{valign=b}{\rotatebox{90}{\scriptsize \qquad \quad REPA}} &
      \includegraphics[width=0.15\textwidth]{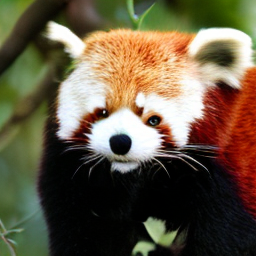} &
      \includegraphics[width=0.15\textwidth]{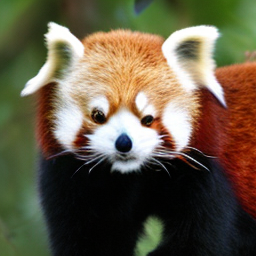} &
      \includegraphics[width=0.15\textwidth]{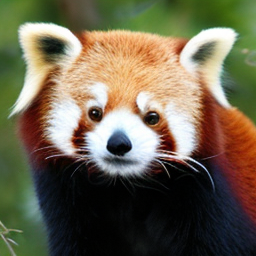} &
      \includegraphics[width=0.15\textwidth]{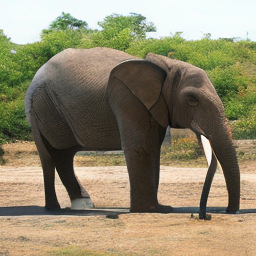} &
      \includegraphics[width=0.15\textwidth]{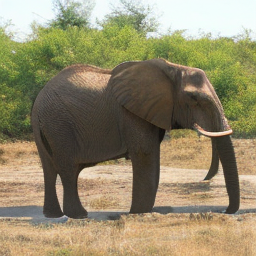} &
      \includegraphics[width=0.15\textwidth]{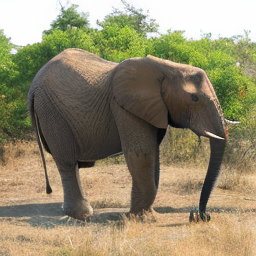} \\
      
      \adjustbox{valign=b}{\rotatebox{90}{\scriptsize \qquad \quad AHPA}} &
      \includegraphics[width=0.15\textwidth]{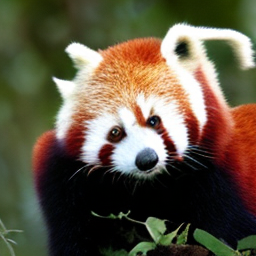} &
      \includegraphics[width=0.15\textwidth]{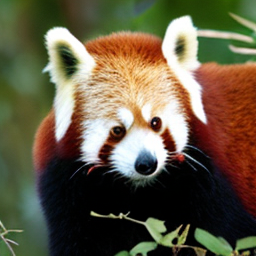} &
      \includegraphics[width=0.15\textwidth]{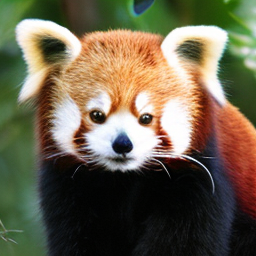} &
      \includegraphics[width=0.15\textwidth]{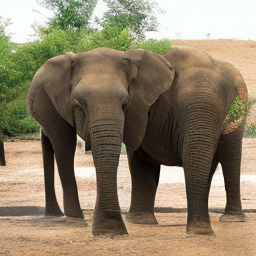} &
      \includegraphics[width=0.15\textwidth]{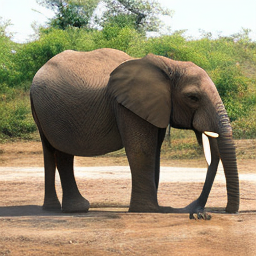} &
      \includegraphics[width=0.15\textwidth]{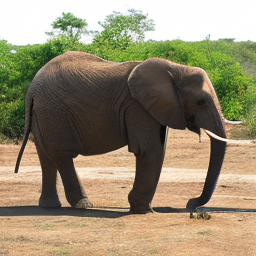} \\
      
      \addlinespace[4pt] 

      
      \adjustbox{valign=b}{\rotatebox{90}{\scriptsize \qquad \quad SRA 2}} &
      \includegraphics[width=0.15\textwidth]{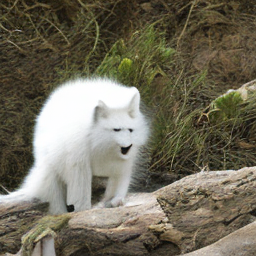} &
      \includegraphics[width=0.15\textwidth]{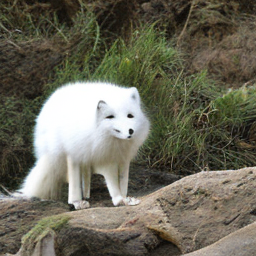} &
      \includegraphics[width=0.15\textwidth]{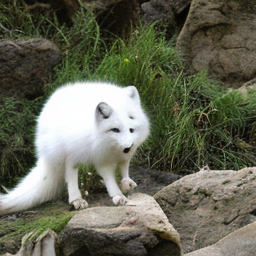} &
      \includegraphics[width=0.15\textwidth]{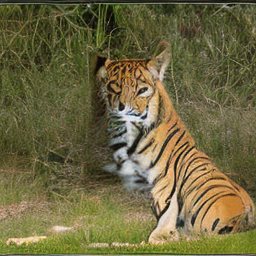} &
      \includegraphics[width=0.15\textwidth]{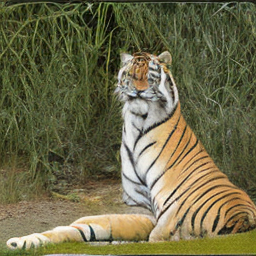} &
      \includegraphics[width=0.15\textwidth]{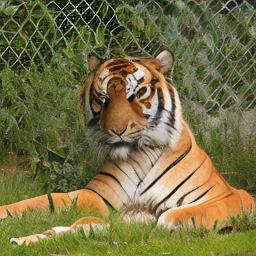} \\
      
      \adjustbox{valign=b}{\rotatebox{90}{\scriptsize  \qquad \quad REPA}} &
      \includegraphics[width=0.15\textwidth]{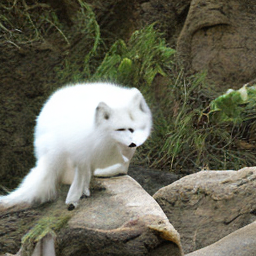} &
      \includegraphics[width=0.15\textwidth]{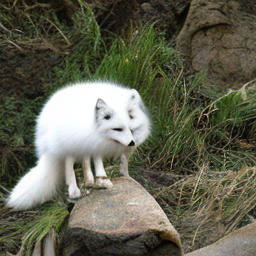} &
      \includegraphics[width=0.15\textwidth]{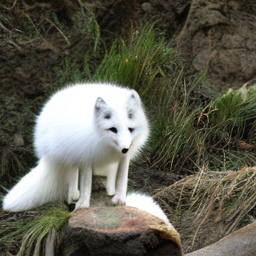} &
      \includegraphics[width=0.15\textwidth]{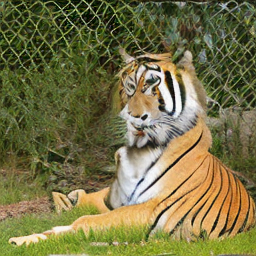} &
      \includegraphics[width=0.15\textwidth]{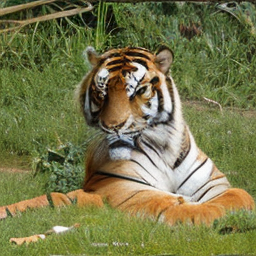} &
      \includegraphics[width=0.15\textwidth]{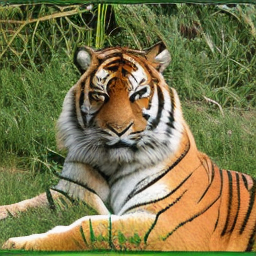} \\
      
      \adjustbox{valign=b}{\rotatebox{90}{\scriptsize \qquad \quad AHPA}} &
      \includegraphics[width=0.15\textwidth]{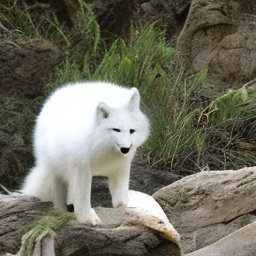} &
      \includegraphics[width=0.15\textwidth]{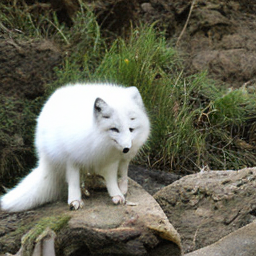} &
      \includegraphics[width=0.15\textwidth]{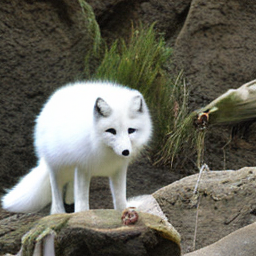} &
      \includegraphics[width=0.15\textwidth]{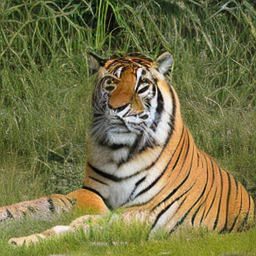} &
      \includegraphics[width=0.15\textwidth]{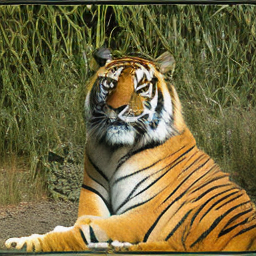} &
      \includegraphics[width=0.15\textwidth]{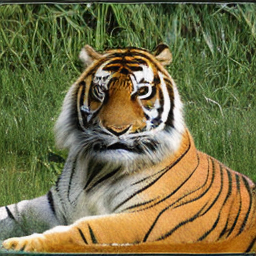} \\
    \end{tabular}
  }

  \caption{\textbf{Qualitative comparison of generative quality on ImageNet $256\times256$.} We visualize the evolution of samples across three methods—SRA 2, REPA, and our AHPA—all utilizing the SiT-XL backbone. To ensure a rigorous controlled comparison, all samples are generated using a fixed classifier-free guidance scale (CFG=4.0) with identical initial noise latents, random seeds, and class labels.}
  \label{fig:qualitative_large}
\end{figure*}

To qualitatively evaluate the acceleration efficiency, we visualize the generated samples at various training milestones (100K, 200K, and 400K) in Figure~\ref{fig:qualitative_large}. The results highlight the distinct advantages of AHPA in terms of both pixel-level fidelity (FID) and structural coherence (sFID). While static alignment baselines like SRA 2 and REPA often struggle with granularity mismatch—resulting in blurred textures or fragmented geometries in mid-training stages—AHPA successfully maintains directional consistency across the generative manifold. Specifically, AHPA produces sharper object boundaries and more faithful semantic patterns (e.g., animal fur and complex structures) much earlier than the baselines. This visual evidence confirms that our adaptive hierarchical prior effectively guides the model to transition from global anchoring to local refinement, leading to superior generative performance under limited computational budgets.
\clearpage

\section{Detailed Training Cost Comparison}
\label{app:detailed_cost_comparison}
Here we analyze the training computational overhead of AHPA. To ensure a fair and efficient comparison, we pre-compute and store the VAE hierarchical features offline, thereby eliminating real-time encoding latency during training. 

As summarized in Table~\ref{tab:detailed_cost_comparison}, AHPA demonstrates superior computational efficiency compared to state-of-the-art alignment-based methods. While methods like SRA~\cite{jiang2026sra} or REPA~\cite{yu2024repa} incur significant overhead (up to +73\% GFLOPs and +71\% latency) by relying on heavy external teacher models (e.g., DINOv2 or CLIP), AHPA is external-free. By utilizing a streamlined single-layer 3$\times$3 convolution for alignment and a lightweight 5-layer MLP for dynamic routing, AHPA adds only a negligible \textbf{5\%} to the total GFLOPs.

Even when accounting for the channel projection layers and the routing mechanism, AHPA maintains a minimal parametric footprint of only 2.5M additional parameters. Our method achieves a high training throughput of 7.4 Iter/s, which is nearly on par with the vanilla SiT-XL/2 baseline (8.1 Iter/s) and significantly faster than external-guided counterparts. These results underscore that AHPA provides a highly frugal yet effective path for feature alignment.

\begin{table}[ht]
\centering
\caption{\textbf{Detailed Training Computational Cost Comparison on ImageNet $256\times256$}. We evaluate AHPA against various alignment baselines using SiT-XL/2 as the transformer backbone. \#EFP is decomposed into \textit{External Model Params + Alignment Head Params}. All benchmarks are conducted on NVIDIA A100 GPUs with a fixed batch size of 256.}
\label{tab:detailed_cost_comparison}
\vspace{0.5em}
\normalsize 
\renewcommand{\arraystretch}{1.2} 
\setlength{\tabcolsep}{12pt} 
\begin{tabular}{l c c c c}
\toprule
\textbf{Method} & \textbf{\#EFP (M)} & \textbf{TS (Iter/s)} $\uparrow$ & \textbf{GFLOPs} $\downarrow$ & \textbf{Latency (ms)} $\downarrow$ \\
\midrule
SiT-XL/2 (Baseline) & $0 + 0$ &8.1 & 114.46 & 18.05 \\
+ REPA \cite{yu2024repa} & $86 + 8$ & 6.3 {\color{red}\scriptsize (-22\%)} & 138.50 {\color{red}\scriptsize (+21\%)} & 22.81 {\color{red}\scriptsize (+26\%)} \\
+ SRA \cite{jiang2026sra} & $481 + 2$ & 5.1 {\color{red}\scriptsize (-37\%)} & 197.57 {\color{red}\scriptsize (+73\%)} & 30.90 {\color{red}\scriptsize (+71\%)} \\
+ SRA 2 \cite{wang2026sra2variationalautoencoder} & $0 + 18$ & 7.2 {\color{red}\scriptsize (-11\%)} & 118.55 {\color{red}\scriptsize (+4\%)} & 19.20 {\color{red}\scriptsize (+6\%)} \\
\rowcolor{highlightpink} \cellcolor{white}\textbf{+ AHPA (Ours)} & \textbf{0 + 2.5} & \textbf{7.4} {\color{red}\scriptsize (-9\%)} & \textbf{119.85} {\color{red}\scriptsize (+5\%)} & \textbf{19.80} {\color{red}\scriptsize (+10\%)} \\
\bottomrule
\end{tabular}
\end{table}

\section{Limitations}
\label{app:limitations}

While AHPA demonstrates significant improvements in training efficiency and generative quality, we acknowledge several limitations that provide avenues for future work:

\begin{itemize}
    \item \textbf{Dependence on Pre-trained VAEs:} Our method relies on the hierarchical feature space of a frozen VAE (e.g., SD-VAE). While these features are robust for general image synthesis, the effectiveness of AHPA might be constrained if the pre-trained encoder fails to capture domain-specific nuances in highly specialized datasets (e.g., medical imaging or high-resolution remote sensing).
    
    \item \textbf{Heuristic Layer Partitioning as a Principled Baseline:} While our group-wise PCA and diagnostic analysis (Sec. ~\ref{sec:motivation}) establish a robust empirical foundation for partitioning VAE layers into functional groups (G1--G4), these boundaries are currently predefined to maintain architectural simplicity and transparency. By fixing these groups, we demonstrate that AHPA can achieve significant gains using minimalist inductive biases without the need for complex, end-to-end grouping protocols. Nevertheless, developing a fully automated, learnable partitioning mechanism represents an exciting future direction that could further enhance AHPA's adaptability to heterogeneous encoder architectures.
    
    \item \textbf{Focus on Class-Conditional Synthesis:} This study primarily focuses on class-conditional ImageNet synthesis and MS-COCO text-to-image tasks. The generalization of adaptive hierarchical alignment to other modalities, such as video generation or 3D synthesis, requires further investigation to account for temporal and volumetric prior requirements.
    
    \item \textbf{Fixed Number of Groups:} In our experiments, we utilize four functional groups. While this granularity is sufficient for existing DiT/SiT scales, extremely large models (e.g., $>$10B parameters) might benefit from an even finer-grained hierarchical schedule which we have not yet explored.
\end{itemize}

\section{Broader Impacts}
\label{app:broader_impacts}

Our work on Adaptive Hierarchical Prior Alignment (AHPA) has several societal implications, ranging from environmental benefits to ethical considerations in generative AI.

\paragraph{Positive Societal Impacts.}
The most direct positive impact is the significant reduction in the environmental footprint of training large-scale generative models. By achieving a 17$\times$ acceleration in convergence, AHPA substantially lowers the total energy consumption and associated carbon emissions required to reach state-of-the-art performance. Furthermore, our "frugal" framework—which avoids reliance on massive external teacher models—promotes the democratization of AI research. It enables researchers with limited computational budgets to develop high-quality generative systems without needing the prohibitive hardware resources typically required for traditional alignment methods.

\paragraph{Negative Societal Impacts and Mitigations.}
As AHPA enhances the visual fidelity and structural consistency of synthetic images, it inherently increases the risk of misuse for creating deepfakes or disinformation. High-quality synthetic content can be weaponized to deceive the public or impersonate individuals. To mitigate these risks, we advocate for the integration of digital watermarking (e.g., StegaStamp) and the development of robust forgery detection algorithms alongside the release of such models. Additionally, since AHPA aligns with VAE features trained on large-scale datasets (e.g., ImageNet), it may inadvertently inherit or amplify societal biases present in the training data. We encourage users of our framework to perform bias audits and apply post-generation filtering when deploying these models in sensitive applications.

\paragraph{Ethical Statement.}
This research does not involve human subjects, personal data, or harmful biological/chemical agents. We have conducted our experiments in accordance with established ethical guidelines for machine learning research.
\section{Computational Resources and Compute Estimates}
\label{app:compute_resources}

In accordance with the NeurIPS 2026 guidelines, we provide a detailed disclosure of the computational resources and total compute investment utilized for this research.

\subsection{Hardware and Environment}
All experiments, including preliminary explorations and final evaluations, were performed on an internal high-performance computing cluster. The specific hardware configuration per node is as follows:
\begin{itemize}
    \item \textbf{Compute Workers:} 8 $\times$ NVIDIA A100 GPUs (80GB SXM5).
    \item \textbf{CPU:} Dual AMD EPYC 7742 64-Core Processors.
    \item \textbf{Memory:} 1 TB DDR4 system RAM.
    \item \textbf{Storage:} 4 TB NVMe local scratch space.
    \item \textbf{Software:} CUDA 12.1, PyTorch 2.2, and NCCL 2.18.
\end{itemize}

\subsection{Individual Run Compute Requirements}
We report the estimated wall-clock time and GPU hours for single successful training runs (400K iterations) on ImageNet $256\times256$. The specific costs are:
\begin{itemize}
    \item \textbf{SiT-B/2 + AHPA:} $\sim$44 hours on 8 GPUs ($\approx$ 352 GPU hours).
    \item \textbf{SiT-L/2 + AHPA:} $\sim$78 hours on 8 GPUs ($\approx$ 624 GPU hours).
    \item \textbf{SiT-XL/2 + AHPA:} $\sim$112 hours on 8 GPUs ($\approx$ 896 GPU hours).
    \item \textbf{Vanilla Baselines (Re-trained):} Totaling $\sim$1,800 GPU hours to ensure fair comparison under identical hardware.
    \item \textbf{MS-COCO Text-to-Image:} $\sim$65 hours on 16 GPUs ($\approx$ 1,040 GPU hours).
\end{itemize}

\subsection{Total Project Compute Disclosure}
The full research project required significantly more compute than the final experiments reported in the main paper to account for the iterative development process:
\begin{itemize}
    \item \textbf{Preliminary Explorations:} Initial diagnostic probing of VAE hierarchical groups and verification of the non-stationary requirement ($\sim$500 GPU hours).
    \item \textbf{Hyperparameter Tuning:} Searching for the optimal Dynamic Router architecture and alignment weight $\lambda$ ($\sim$1,200 GPU hours).
    \item \textbf{Failed Experiments:} Early attempts at content-adaptive routing and alignment at late-stage blocks that did not yield performance gains ($\sim$800 GPU hours).
\end{itemize}

\textbf{Total Estimated Compute:} We estimate the total computational investment for the entire research cycle to be approximately 7,200 NVIDIA A100 GPU hours.

\section{More Related Works}
\paragraph{Generative Models and Diffusion Transformers.}
Image generation has witnessed a transition from pixel-space denoising models, such as DDPM~\cite{ho2020ddpm} and DDIM~\cite{song2021denoising}, to Latent Diffusion Models (LDMs)~\cite{rombach2022LDM-4} that operate in compressed latent spaces. Architecturally, while early frameworks relied on U-Net backbones~\cite{ho2020ddpm,rombach2022LDM-4}, modern paradigms have shifted towards Transformer-based architectures like DiT~\cite{peebles2023scalable} and SiT~\cite{ma2024sit}, which leverage self-attention mechanisms for superior spatial modeling~\cite{peebles2023scalable}. Despite these advances, diffusion transformers typically require extensive training iterations to achieve convergence~\cite{singh2025irepa,wu2025reg}. While acceleration can be achieved through structural modifications like masked training~\cite{zheng2024maskdit}, these methods often necessitate changes to the training paradigm.

\end{document}